\pdfoutput=1

\documentclass[11pt]{article}

\usepackage{authblk}
\usepackage[]{acl}

\usepackage{graphicx}
\usepackage{subfigure} 
\usepackage{amsmath}
\usepackage{adjustbox} 
\usepackage{amsthm}
\usepackage{booktabs}
\usepackage{algorithm}
\usepackage{algpseudocode}
\usepackage{amssymb}
\usepackage{multirow}
\usepackage{longtable}
\usepackage{comment}
\usepackage[normalem]{ulem}
\usepackage[utf8]{inputenc}
\usepackage{csquotes}
\usepackage{flushend}

\usepackage{times}
\usepackage{latexsym}

\usepackage[T1]{fontenc}

\usepackage{microtype}
\usepackage{xspace}

\usepackage{authblk}

\usepackage{todonotes}
\usepackage{xcolor}

\newcommand{\method}{GLUE-X\xspace}

\title{GLUE-X: Evaluating Natural Language Understanding Models from an Out-of-Distribution Generalization Perspective}

\author{
    \textbf{Linyi Yang}$^{1,2}$\thanks{\ \ Equal contribution. Random order. Shuibai Zhang finished this work at Westlake University as an intern.}, \textbf{Shuibai Zhang}$^{1,4,*}$, \textbf{Libo Qin}$^{3}$, \textbf{Yafu Li}$^{1}$, \textbf{Yidong Wang}$^{1}$, \textbf{Hanmeng Liu}$^{1}$, \\
    \textbf{Jindong Wang}$^{5}$, \textbf{Xing Xie}$^{5}$, \textbf{Yue Zhang}$^{1,2}$ \thanks{\texttt{\{yanglinyi, zhangyue\}@westlake.edu.cn} Yue Zhang is the corresponding author.}
} 
\affil{$^{1}$ School of Engineering, Westlake University\\
    $^{2}$ Institute of Advanced Technology, Westlake Institute for Advanced Study\\
    $^{3}$ School of Computer Science and Engineering, Central South University\\
    $^{4}$ University of Electronic Science and Technology of China\\
    $^{5}$ Microsoft Research Asia\\
    \texttt{}\\
    
}
\date{}    

\begin{document}

\maketitle

\begin{abstract}
Pre-trained language models (PLMs) are known to improve the generalization performance of natural language understanding models by leveraging large amounts of data during the pre-training phase. However, the out-of-distribution (OOD) generalization problem remains a challenge in many NLP tasks, limiting the real-world deployment of these methods. This paper presents the first attempt at creating a unified benchmark named \method for evaluating OOD robustness in NLP models, highlighting the importance of OOD robustness and providing insights on how to measure the robustness of a model and how to improve it. The benchmark includes 15 publicly available datasets for OOD testing, and evaluations are conducted on 8 classic NLP tasks over 21 popularly used PLMs. Our findings confirm the need for improved OOD accuracy in NLP tasks, as significant performance degradation was observed in all settings compared to in-distribution (ID) accuracy.\looseness=-1
\end{abstract}

\section{Introduction}

Pre-trained Language Models (PLMs) \cite{qiu2020pre,bommasani2021opportunities} have achieved competitive performance across standard NLP benchmarks \cite{blasi22acl}, such as GLUE \cite{wang2018glue} and SuperGLUE \cite{wang2019superglue}. However, recent studies \cite{gururangan2018annotation,ribeiro2019red,kaushik2020learning,ribeiro2020beyond,ruder2021challenges} show concerns that models are yet not close to achieving proper natural language understanding, essential questions being raised about their robustness \cite{srivastava2020robustness,wang2021measure} and underlying sensitivity to systematic biases \cite{probing2019acl,sagawa2020investigation}.
Such issues manifest in the performance decay, especially for out-of-distribution (OOD) generalization when the test distribution differs from training \cite{arora2021types,malinin2021shifts,hupkes2022state}. 

OOD generalization has been systemically studied for Computer vision (CV) and artificial general intelligence (AGI) \cite{koh2021wilds,srivastava2022beyond}, for which large evaluation datasets are available. While sharing the same aspirational goal, existing evaluations \cite{kaushik2018much,min2019compositional,gardner2020evaluating} and methods~\cite{hendrycks2020pretrained,bommasani2021opportunities} for OOD generalization of NLP contain only one or a few tasks~\cite{wu2021polyjuice,wang2021robustness,howard2022neurocounterfactuals,lu2022rationale}, which do not adequately capture limitations of existing models, resulting in inflated test accuracy \cite{tu2020empirical,ribeiro2020beyond}. Thus it remains a gap in evaluating models in a unified way by executing a range of text classification tasks.

To facilitate research in this direction, we introduce the \method benchmark for evaluating the out-of-distribution performance of PLMs. \method expands upon previous multi-task benchmarks \cite{fewnlu,xu2020clue,xu2021fewclue} by including test data from multiple domains, covering eight standard tasks in GLUE, with an average 2 test domains for each task, allowing comprehensive cross-distribution evaluations. Specifically, \method focuses on domain generalization, where a model trained on a source domain can be directly generalized to target domains without any labeled or unlabeled data from target domains. It also enables the analysis of two main factors affecting the cross-domain generalization performance, namely the pre-trained language model (e.g., architecture, size, etc.) and different training strategies (e.g., fine-tuning, prompt-tuning \cite{chen2022adaprompt}, linear probing \cite{wu2019understanding}, and domain-generalization training \cite{wang2023robustness}).

Using \method, we evaluate the performance of \emph{21} pre-trained language models in a unified setting and under the same experimental conditions. In addition, we consider \emph{3} tuning strategies designed for improving single-source domain generalization: linear probing \cite{tripuraneni2020theory,wu2019understanding}, fine-tuning, and the linear probing then fine-tuning method (LP-FT) \cite{kumar2022finetuning}. Finally, we analyze the internal causes of OOD robustness at the feature level by measuring the rationale overlap between human and model predictions \cite{lei2016rationalizing}.

Results show that the average accuracy of PLMs on cross-domain evaluations falls significantly short of human performance, even for the highest-performing model (81.3\% -- human versus 74.6\% -- model). In contrast to the GLUE leaderboard, where over 20 single-model results outperform human baselines, none of the backbones included in \method is able to surpass human performance under the same evaluation setting. These findings suggest the importance of cross-distribution evaluation for natural language processing. In addition, evidence shows that the superior performance of PLMs on GLUE may be relatively superficial and less useful as a performance indicator in practice.

Detailed analysis shows that (1) no one backbone can significantly outperform the others across all tasks, which is consistent with the conclusion \cite{wenzel2022assaying} in the computer vision; (2) surprisingly, the influence of model architectures is somehow more significant than the model parameters towards the OOD robustness; (3) the ID and OOD performance holds a linear correlation in most cases for text classifications; (4) in terms of the tuning strategy, we show that linear probing and then fine-tuning can slightly improve the OOD performance compared to standard fine-tuning. 

To our knowledge, we are the \emph{first} to systemically evaluate natural language understanding systems for cross-distribution generalization on genuine data compared to human performance. More importantly, we make datasets of cross-domain evaluations for all typical text classification tasks, which allows us to report OOD results under the same experimental conditions. We open-source the codebase and datasets \footnote{\url{https://github.com/YangLinyi/GLUE-X}}. The GLUE-X leaderboard is available at \url{https://gluexbenchmark.com/}.

\section{Related Work}


\paragraph{Benchmarking Robustness to OOD.}

Recent work~\cite{ibrahim2022robustness} finds that today’s leading PLMs are not robust to changing domains, where some OOD test samples varied during training. In particular, pre-trained transformers can rely heavily on spurious patterns (artefacts) \cite{gururangan2018annotation,kaushik2020learning,tu2020empirical}. For this reason, the standard held-out accuracy can overestimate the performance \cite{sagawa2020investigation,kaushik2021learning}, and evaluating the OOD robustness is crucial for real-world applications, which require models to hold good transferability. Consequently, there is a rising concern about improving dataset and benchmark development. Recent work introduces new out-of-distribution benchmarks for graphs \cite{gui2022good}, optical character recognition (OCR) \cite{larson2022evaluating},  computer vision (CV) \cite{ibrahim2022robustness}, time series tasks \cite{gagnon2022woods}, and artificial general intelligence (AGI) \cite{koh2021wilds,srivastava2022beyond}. However, evaluating the out-of-distribution generalization in a multi-task setting has received relatively little attention for NLP.

There is a line of work focusing on the development of challenge datasets, representing as Adversarial NLI~\cite{nie-etal-2020-adversarial}, Dynabench \cite{kiela2021dynabench}, Contrastive Set \cite{gardner2020evaluating}, and AdvGLUE \cite{advglue} where examples are created to be difficult for current models via an iterative, adversarial, and human-and-model-in-the-loop procedure. However, these datasets focus on robustness and stability issues rather than generalization and the artifact. In contrast, GLUE-X contains both off-the-shelf and self-collected datasets to implement cross-distribution tests.

Prior work \cite{wenzel2022assaying} observed that OOD performance holds a linear correlation with ID accuracy in CV based on 172 publicly available datasets and 31k networks, while their relationship is largely dataset-dependent. However, this conclusion has been found somewhat controversial, as \citet{teney2022id} argue that the selection of datasets influences the OOD performance.

\paragraph{Existing Benchmarks for NLU.}



There have been different types of leaderboards towards evaluating natural language understanding (NLU) systems. Examples of building challenging benchmarks in recent years include GLUE \cite{wang2018glue}, SuperGLUE \cite{wang2019superglue}, FewGLUE \cite{fewglue}, FEVER \cite{petroni2021kilt}, FewNLU \cite{fewnlu}, and AdvGLUE \cite{advglue}. In particular, FewGLUE and FewNLU focus on the few-shot learning challenge. The performance decay of NLP models has been found in real-world deployment because of the arises of the OOD generalization challenge as well as robustness issues, such as adversarial robustness. Similar to our work, other benchmarks, such as AdvGLUE \cite{advglue}, leverage the training set extracted from GLUE for each task. Differently, we consider evaluating OOD performance in a general multi-task setting, where the test data arise from one or more different distributions. 



\paragraph{Domain Generalization} (DG)~\cite{wang2022generalizing} aims to learn a generalized model that is robust to unseen distributions using training data from multiple domains~\cite{balaji2018metareg,dou2019domain,vu2022domain,varshney2022investigating}. We focus on the single-source DG~\cite{huang2020self,krueger2021out,wang2022generalizing} setting, which is a popular setting for measuring the OOD robustness in NLP~\cite{hendrycks2020pretrained}, and aligns with the GLUE leaderboard. As stated by a recent taxonomy and review towards generalisation research in NLP \cite{hupkes2022state}, current work does not provide standardized data or procedures for generalization testing, while we use GLUE-X as the first attempt towards this goal.

\begin{table}[t!]
\centering
\small
\resizebox{\linewidth}{!}{%
\begin{tabular}{l|llr}
\toprule
\textbf{Task} & \textbf{ID} & \textbf{OOD} & \textbf{Size} \\ \cmidrule(r){1-4}
\multirow{4}{*}{\begin{tabular}[c]{@{}l@{}}Sentiment \\ Analysis\end{tabular}} & \multirow{4}{*}{SST-2} & IMDB & 50,000 \\
 &   & Yelp  & 598,000 \\
 & & Amazon & 4,000,000 \\
 & & Flipkart & 205,041 \\ \cmidrule(r){1-4}
\begin{tabular}[c]{@{}l@{}}Linguistic \\ Acceptability\end{tabular}                       & CoLA                                                                      & \begin{tabular}[c]{@{}l@{}}Grammar \\ Test\end{tabular} & 304,277 \\ \cmidrule(r){1-4}
\begin{tabular}[c]{@{}l@{}}Textual \\ Similarity\end{tabular}                             & STSB                                                                      & SICK                                                    & 9,840 \\ \cmidrule(r){1-4}
\multirow{3}{*}{\begin{tabular}[c]{@{}l@{}}Natural \\ Language \\ Inference\end{tabular}} & \multirow{3}{*}{\begin{tabular}[c]{@{}l@{}}MNLI\\ (matched)\end{tabular}} & MNLI\-mis                                               & 9,832 \\
&  & SNLI & 570,152 \\
&  & SICK & 9,840 \\ \cmidrule(r){1-4}
\begin{tabular}[c]{@{}l@{}}Question\\Answering \\NLI\end{tabular}                             & QNLI                                                                      & \begin{tabular}[c]{@{}l@{}}NewsQA\\(Reconstructed)\end{tabular}                                                      & 119,525 \\ \cmidrule(r){1-4}
\multirow{2}{*}{Textual Entailment} & \multirow{2}{*}{RTE} & SciTail & 26,527 \\
 & & HANS & 60,000 \\ \cmidrule(r){1-4}
\multirow{4}{*}{\begin{tabular}[c]{@{}l@{}}Paraphrase\end{tabular}} & \multirow{2}{*}{MRPC} & QQP & 404,276  \\
&  & Twitter & 16,777 \\ \cmidrule(r){2-4} & \multirow{2}{*}{QQP} & MRPC & 4,076  \\
 & & Twitter & 16,777 \\
 \bottomrule 
\end{tabular}
}
\caption{Data statistic of GLUE-X, which describes the source and size for OOD tests over different tasks.}
\label{tab:data_statistic}
\end{table}

\section{Data and Settings}

The goal of \method is to provide an extension of GLUE with the same training data but multifarious OOD test sets.

\subsection{Overview of \method}

The evaluation in \method is intrinsically related to the domain generalization task considering a practical and challenging setting, where a model trained on multiple source domains can be directly generalized to a target domain without any labeled or unlabeled data from
the target domain \cite{muandet2013domain}.  We articulate the following tasks and datasets in \method.

\noindent\textbf{Tasks.} As a benchmark styled after GLUE \cite{wang2018glue}, we consider eight tasks in \method: Sentiment Analysis (\emph{SST-2}), Natural Language Inference (\emph{MNLI, QNLI}), Textual Entailment (\emph{RTE}), Paraphrase (\emph{MRPC, QQP}), Textual Similarity (\emph{STS-B}) and Linguistic Acceptability (\emph{CoLA}). \footnote{The WNLI task is not included in \method since there is no sufficient in-domain data for constructing OOD tests~\cite{wang2022pre,xlnet,yang2022factmix}.}

\noindent\textbf{Datasets.} \method follows the same in-domain training data and evaluation metrics as GLUE \cite{wang2018glue}. To construct the out-of-domain test, we adopt popular datasets extracted from different domains while keeping the same prediction labels as the original tasks in GLUE. The detailed data statistics are shown in Table~\ref{tab:data_statistic}. 

\subsection{Dataset Curation}

We construct test sets for each task under the requirement that they share the same label types with the training set. To this end, \method contains 15 OOD datasets, including publicly available datasets (Amazon, HANS, etc) and newly collected/re-constructed datasets (Grammar Test). In particular, we select the OOD datasets for each task, including sentiment analysis -- IMDB \cite{maas-etal-2011-learning}, Yelp \cite{zhang2015character}, Amazon \cite{kaushik2020learning}, and Flipkart \cite{flipkart_2023}; linguistic acceptability -- Grammar Test; textual similarity -- SICK \cite{zhang2018multi}; NLI -- MNLI-Mismatched \cite{williams2017broad}, SNLI \cite{bowman2015large}, and SICK \cite{zhang2018multi}; Textual Entailment -- RTE; Paraphrase -- MRPC and QQP \cite{bentivogli2009fifth,dolan-2005-automatically,wang2017bilateral,mccoy2019right}. Regarding the QNLI task, we convert instances from NewsQA \cite{newsqa} to the consistent data format of QNLI for conducting the OOD evaluation. The detailed description of the newly collected dataset, Grammar Test, can be found in Appendix \ref{sec:appendixA}.

SICK contains multiple labels, including textual similarity, also used as an OOD test set of the textual similarity task. We rounded floating number labels of textual similarity to integers from 0 to 4, converting it into a five-class dataset to align with other classification tasks in GLUE-X. In addition, MRPC and QQP are leveraged as OOD datasets of each other as the paraphrasing task. 

\subsection{Metrics}

We first average metrics to get a score for those tasks with multiple metrics. Following GLUE and SuperGLUE, we then report the score of NLU models by averaging the scores of all tasks as the OOD performance.  For rankings, in addition to the robustness rank by considering the decreased ratio between OOD and ID performance, we adopt Friedman rank \cite{friedman1940comparison} over multiple tasks:
\begin{equation*}
\operatorname{rank}_{F}  = \frac{1}{n} \sum_{i=1}^{n} \operatorname{rank}_i,
\end{equation*}
where $n$ is the number of tasks (e.g., $n=8$ in Table~\ref{tab:overall_per}) and $\operatorname{rank}_i$ is the rank of the performance in the $i$-th task considering in the GLUE-X. We report the robustness ranking in terms of the decreased ratio of OOD performance and Friedman rank.

\subsection{Post-hoc Analysis}

In addition to quantitative analysis, we choose two tasks, sentiment analysis, and natural language inference, for post-hoc feature analysis \cite{lei2016rationalizing}. We adopt the sensitivity of contextual decomposition technique \cite{jin2019towards,yang-etal-2021-exploring}, which removes part of inputs from the sequence text to evaluate a model's sensitivity to them, thereby allowing for identifying important features. The output is the overlap between rationales by models and humans, which to some extent represents the trust of models \cite{jacovi2020towards,yang-etal-2020-generating}. 

Formally, given a phrase \emph{p} starting with the negative limitations in the \emph{k-th} document $\mathcal{D}^{(k)}$, we sample the documents which contain the same phrase \emph{p} to alleviate the influence by chance when there are multiple shreds of evidence saturating the prediction. The window size of the phrase \emph{p} is limited to 3. Taking sentiment analysis for example, given \textit{``This movie was so unbelievably bad''} if we only remove the non-causal word \textit{movie}, the prediction is not expected to change for a robust model.

The importance score is computed as follows:
\begin{equation*}
\small
\phi(\mathbf{p}, \widehat{\mathcal{D}^{(k)}})=\mathbb{E}_{\widehat{\mathcal{D}^{(\beta)}}}\left[l\left(\widehat{\mathcal{D}^{(\beta)}}; \widehat{\mathcal{D}}\right)-l\left(\widehat{\mathcal{D}^{(\beta)}} \backslash \mathbf{p} ; \widehat{\mathcal{D}}\right)\right],
\end{equation*}
where \(\mathcal{D}^{(\beta)}\) denotes the resulting text after masking out a single token (phrase) starting with the negative pronoun (un-, non-, etc.) in the length of \(N\) surrounding the phrase \(\mathbf{p}\). We use \(l\left(\widehat{\mathcal{D}^{(\beta)}} \backslash \mathbf{p}; \widehat{\mathcal{D}}\right)\) to represent the model prediction logits of the ground-truth class after replacing the masked-out context. \(\backslash \mathbf{p}\) indicates the operation of masking out the phrase \(p\) in a given document. 

\begin{table}[t]
\centering
\small
\begin{tabular}{l|ccc}
\toprule
\begin{tabular}[c]{@{}l@{}}Type\\ Backbone \end{tabular} & \begin{tabular}[c]{@{}c@{}}Training\\ (GPU Hours)\end{tabular} & \begin{tabular}[c]{@{}c@{}}Inference\\ (GPU Hours)\end{tabular} & \begin{tabular}[c]{@{}c@{}}Total\\ Hours\end{tabular} \\ \midrule
\begin{tabular}[c]{@{}l@{}}BERT-large\end{tabular}& 440 & 240 & 680  \\
\begin{tabular}[c]{@{}l@{}}T5-large\end{tabular}& 792 & 420 & 1,212  \\
\begin{tabular}[c]{@{}l@{}}ALBERT-base\end{tabular} & 165 & 120 & 285   \\ \bottomrule
\end{tabular}
\caption{The training and testing cost of GLUE-X.}
\label{tab:cost}
\end{table}

\subsection{Models and Training Strategies}
\noindent\textbf{Models.} To ensure that our results are relevant for both researchers and practitioners, we consider both top-performing model backbones and cost-efficient methods: \emph{Discriminative Models} -- BERT-base, BERT-large \cite{bert}, RoBERTa-base, RoBERTa-large \cite{roberta}, XLNet-base, XLNet-large \cite{xlnet}; \emph{Generative Models} -- BART-base, BART-large \cite{bart}, T5-small, T5-base, T5-large \cite{t5}, GPT2, GPT2-medium, GPT2-large \cite{gpt2}; \emph{Generative and Discriminative Models} -- ELECTRA-small, ELECTRA-base, ELECTRA-large \cite{electra}; \emph{Cost-Efficient Models} -- ALBERT-base \cite{albert}, and DistilBERT-base \cite{distilbert}. We also report the results of GPT-3 \cite{gpt3} and GPT-3.5 \cite{openai2023gpt4} through in-context learning. We follow the official implementations of several pre-trained language models from Huggingface\footnote{\url{https://huggingface.co/models}} to reproduce results on GLUE using the validation set and test these models on \method. The hyper-parameters of each model are selected by using grid search and can be found in Appendix. 

\noindent\textbf{Fine-tuning Strategies.} We investigate the efficacy of different fine-tuning strategies for OOD generalization. In particular, we consider three paradigms: standard fine-tuning, fine-tuning only the head (linear probing), and linear probing then fine-tuning. The detailed training cost and inference speed estimated by a single V100 are shown in Table \ref{tab:cost}, in which we evaluate the performance using the in- and out-of-domain test data, recording the training cost in GLUE and \method. We use 50 NVIDIA Tesla V100 GPU cards and 8 NVIDIA A100 GPU cards and spend 10,000+ GPU hours based on the estimation with a single V100 card.

\begin{table*}[t]
\centering
\begin{adjustbox}{width=\textwidth}
\begin{tabular}{lrrrrrrr}
\toprule
\multirow{2}{*}{\textbf{Pre-trained Models}}                 & \textbf{Avg}                    & \textbf{Avg}         & \textbf{Avg}         & \textbf{F-Rank}     & \textbf{F-Rank}     & \textbf{Rank} & \textbf{PARAM} \\
                                                    &
\method & GLUE        & $\Delta$↓   & OOD        & ID      & $\Delta$↓        & (M)          \\ \hline
ELECTRA-large \cite{electra}       & \textbf{74.62}            & \textbf{89.18} & \textbf{16.33}                           & \textbf{2.13} & \textbf{2.25} & \textbf{1}    & 334.09                         \\
T5-large \cite{t5}                 & 72.81                                      & 87.70                           & 16.98                           & 2.38                           & 3.00                           & 2                              & 737.67                         \\
RoBERTa-large \cite{roberta}       & 71.62                                      & 87.83                           & 18.46                           & 4.00                           & 3.00                           & 3                              & 355.36                         \\
BART-large \cite{bart}             & 70.38                                      & 87.05                           & 19.15                           & 5.00                           & 3.63                           & 6                              & 406.29                         \\
T5-base \cite{t5}                  & 70.05                                      & 85.92                           & 18.47                           & 5.88                           & 6.13                           & 4                              & 222.90                         \\
XLNet-large \cite{xlnet}           & 69.69                                      & 86.75                           & 19.67                           & 6.00                           & 4.63                           & 8                              & 360.27                         \\
RoBERTa-base \cite{roberta}        & 68.73                                      & 85.27                           & 19.40                           & 7.00                           & 6.63                           & 7                              & 124.65                         \\
ELECTRA-base  \cite{electra}       & 67.78                                      & 85.92                           & 21.11                           & 9.63                           & 8.63                           & 15                             & 108.89                         \\
GPT2-large  \cite{gpt2}            & 66.46                                      & 83.57                           & 20.47                           & 10.88                          & 11.50                          & 10                             & \textbf{774.03}                         \\
BART-base  \cite{bart}             & 65.89                                      & 83.04                           & 20.65                           & 11.00                          & 11.00                          & 12                             & 139.42                         \\
BERT-large  \cite{bert}            & 65.80                                      & 83.26                           & 20.97                           & 11.38                          & 10.38                          & 14                             & 335.14                         \\
T5-small  \cite{t5}                & 65.43                                      & 80.35                           & 18.57 & 12.63                          & 15.00                          & 5                              & 60.51                          \\
ALBERT-base  \cite{albert}         & 65.30                                      & 82.58                           & 20.93                           & 12.88                          & 13.25                          & 13                             & 11.68                          \\
ELECTRA-small  \cite{electra}      & 65.06                                      & 81.50                           & 20.17                           & 13.88                          & 16.13                          & 9                              & 13.48                          \\
GPT2-medium  \cite{gpt2}           & 65.03                                      & 81.84                           & 20.54                           & 12.88                          & 13.63                          & 11                             & 354.82                         \\
XLNet-base  \cite{xlnet}           & 64.57                                      & 82.26                           & 21.50                           & 12.75                          & 12.13                          & 16                             & 116.72                         \\
BERT-base  \cite{bert}             & 64.10                                      & 82.08                           & 21.91                           & 13.88                          & 13.88                          & 17                             & 109.48                         \\
DistilBERT-base  \cite{distilbert} & 61.94                                      & 80.21                           & 22.78                           & 17.75                          & 17.38                          & 18                             & 66.36                          \\
GPT2  \cite{gpt2}                  & 61.16                                      & 79.30                           & 22.88                           & 18.13                          & 17.88                          & 19                             & 124.44                         \\ \bottomrule
\end{tabular}
\end{adjustbox}
\caption{Overall performance sorted by the \method performance. The average accuracy shown in the table is the mean average score of the OOD performance for each task. The average \textbf{$\Delta$↓} indicates the decreased ratio from the average ID accuracy to OOD accuracy. We also provide the Friedman rank \cite{friedman1940comparison} for OOD and ID tests (shown as F-Rank). The robustness rank is sorted by the average ratio of performance decay in ascending order.} 
\label{tab:overall_per}
\end{table*}

\section{Experiments}

We explore the facets of OOD generalization in NLP using \method, highlighting discrepancies to previous findings and discussing their implications.

\subsection{Human Annotation}
We employ human annotators to give predictions on OOD datasets and identify rationales.

\noindent\textbf{Predictions.} 
We use a crowd-sourcing company to recruit editors and annotators to give predictions on 15 OOD datasets. To fairly compare human performance with models, we simulate the models' OOD testing during the manual annotation process. Specifically, annotators are given essential instructions and a few examples from the in-domain dataset that gently guide them to annotate. Then they are asked to label instances from unseen OOD datasets, typically collected from other domains. 1,000 testing samples are used to obtain the human performance for each OOD dataset.

We employ multiple labelers to annotate the same data point (1,000 samples for each dataset) during the annotation to ensure the high quality of the crowdsourcing work. All annotators have an undergraduate degree in English or a PhD in an English-speaking country. In particular, we employ ten people to annotate the SICK dataset as same as the original data \cite{zhang2018multi}. We employ two annotators for labeling the same instance for the other datasets. After the trial phase of data annotation, we set the Inter-Annotator Agreement (IAA) score threshold for each task depending on the difficulty level. Finally, the average IAA over the 15 OOD datasets is 0.857, indicating acceptable agreement.

\begin{table*}[t]
\centering
\small
\begin{adjustbox}{width=\textwidth}
\begin{tabular}{lcccccccc|cc}
\toprule
\multirow{2}{*}{\textbf{Model}} & \multicolumn{1}{c}{\textbf{SST-2}} & \multicolumn{1}{c}{\textbf{MNLI}} & \multicolumn{1}{c}{\textbf{QNLI}} & \multicolumn{1}{c}{\textbf{RTE}} & \multicolumn{1}{c}{\textbf{MRPC}} & \multicolumn{1}{c}{\textbf{QQP}} & \multicolumn{1}{c}{\textbf{STS-B}} & \multicolumn{1}{c}{\textbf{CoLA}} & \multicolumn{1}{c}{\textbf{Avg}} & \textbf{Avg}       \\
                       & \multicolumn{1}{c}{OOD}   & \multicolumn{1}{c}{OOD}  & \multicolumn{1}{c}{OOD} & \multicolumn{1}{c}{OOD}  & \multicolumn{1}{c}{OOD} & \multicolumn{1}{c}{OOD}   & \multicolumn{1}{c}{OOD}  & \multicolumn{1}{c}{OOD} & \multicolumn{1}{c}{OOD} & $\Delta$↓ 
                       \\ \midrule
Human Performance & \textit{92.36} & \textit{84.13} & \textit{81.10} & \textit{86.53} & \textit{79.31} & \textit{78.46} & \textit{80.28} & \textit{58.98} & \textit{80.14} & \textit{7.82}
                       \\ \midrule
ELECTRA-large          & \textbf{94.67}                     & 76.94                    & 80.44                   & 78.74                    & 69.96                   & \textbf{77.24}                     & \textbf{81.14}                   & \textbf{37.85}        & 74.62                   & 16.33                                             \\
T5-large               & 93.83                     & 76.36                    & \textbf{81.72}                   & \textbf{81.52} & \textbf{72.66}       & 72.26                     & 77.86                    & 26.30                    & 72.81                   & 16.98                                             \\
RoBERTa-large          & 93.07                     & 77.28                    & 79.67                   & 77.84                    & 65.19                   & 76.11             & 77.91                    & 25.90                    & 71.62                   & 18.46                                             \\
BART-large             & 93.51                     & 76.09                    & 80.45                   & 71.94                    & 70.56                   & 73.41                     & 76.03                    & 21.06                    & 70.38                   & 19.15                                             \\
T5-base                & 93.52                     & 73.76                    & 80.29                   & 73.85                    & 70.90                   & 73.20                     & 74.98                    & 19.94                    & 70.05                   & 18.47 \\
XLNet-large          & 93.75                     & \textbf{77.59}                    & 79.98                   & 76.29                    & 65.07                   & 65.25                     & 76.86                    & 22.76                    & 69.69                   & 19.67                                             \\
RoBERTa-base         & 92.77                     & 74.21                    & 79.55                   & 67.82                    & 66.89                   & 70.90                     & 74.90                     & 22.81                    & 68.73                   & 19.40                                              \\
ELECTRA-base         & 90.41                     & 75.33                    & 78.10                    & 74.00                       & 59.49                   & 66.55                     & 77.10                     & 21.23                    & 67.78                   & 21.11                                             \\
GPT2-large           & 91.60                      & 73.62                    & 75.77                   & 70.54                    & 62.05                   & 70.16                     & 70.61                    & 17.32                    & 66.46                   & 20.47                                             \\
BART-base            & 91.49                     & 74.32                    & 78.77                   & 63.28                    & 66.56                   & 67.36                     & 72.46                    & 12.90                     & 65.89                   & 20.65                                             \\
BERT-large           & 91.47                     & 73.33                    & 78.79                   & 64.97                    & 62.06                   & 69.26                     & 69.76                    & 16.77                    & 65.80                    & 20.97                                             \\
T5-small             & 89.40                      & 70.67                    & 77.44                   & 63.96                    & 71.63                   & 67.24                     & 72.58                    & 10.55                    & 65.43                   & 18.57                                             \\
ALBERT-base          & 89.97                     & 70.95                    & 77.31                   & 65.70                     & 61.08                   & 68.04                     & 72.17                    & 17.18                    & 65.30                    & 20.93                                             \\
ELECTRA-small        & 89.40                      & 70.57                    & 75.02                   & 65.89                    & 59.64                   & 65.14                     & 72.36                    & 22.48                    & 65.06                   & 20.17                                             \\
GPT2-medium          & 91.42                     & 72.82                    & 77.70                    & 66.60                     & 57.75                   & 67.53                     & 69.16                    & 17.26                    & 65.03                   & 20.54                                             \\
XLNet-base           & 91.54                     & 74.75                    & 76.87                   & 63.47                    & 62.34                   & 65.76                     & 68.29                    & 13.53                    & 64.57                   & 21.50                                              \\
BERT-base            & 89.36                     & 70.92                    & 78.31                   & 59.54                    & 61.83                   & 67.49                     & 67.68                    & 17.66                    & 64.10                    & 21.91                                             \\
DistilBERT-base      & 87.06                     & 70.27                    & 74.27                   & 58.76                    & 61.63                   & 64.96                     & 66.18                    & 12.37                    & 61.94                   & 22.78                                             \\
GPT2                 & 82.46                     & 69.67                    & 76.41                   & 60.55                    & 58.90                    & 64.79                     & 66.26                    & 10.22                    & 61.16                   & 22.88                                             \\
\bottomrule
\end{tabular}
\end{adjustbox}
\caption{Detailed OOD performance for each task in \method. Evaluation metrics for each task are the same as GLUE (the average results are reported for those tasks considering two metrics). The best performance is shown in bold. Human evaluation is simulated in a similar OOD setting by receiving instructions from ID samples while predicting data from OOD datasets. The human baseline is shown in italics if it beats the best-performing model.}
\label{tab:detail}
\end{table*}

\noindent\textbf{Rationale Marking.} Following \citet{kaushik2020learning} and \citet{kaushik2021learning}, we use extractive explanations for marking rationales that support classification decisions. Inspired by \citet{kaushik2021learning} and \citet{lertvittayakumjorn2021explanation}, we leverage the rationale marking annotated by humans to compare with rationale selected by models on sentiment analysis and natural language inference (NLI) tasks. We ask two labelers to annotate sampled instances from IMDB, Yelp, and Amazon datasets for the sentiment analysis task. At the outset, annotators were given instructions and examples that gently guided them to annotate rationales. Only adjectives, adverbs, nouns, and verbs were considered rationale candidates. Besides, rationales were required to carry complete semantic information. We sampled 6,000 instances for each dataset randomly. Using F1 score, the IAA for IMDB, Yelp and Amazon are 0.874, 0.871, and 0.840, respectively. For NLI, we use the explanation dataset, e-SNLI \cite{camburu2018snli}, to assert the models' trust. 


\subsection{Prediction Results} 

\textbf{Overall Performance on \method.} We report the average score of different models sorted in descending order representing the overall performance in Table \ref{tab:overall_per}. In addition to the overall performance, we provide the Friedman Rank for in- and out-of-domain results. From Table \ref{tab:overall_per}, we observe that all pre-trained models involved in \method show significant performance decay under the OOD test compared to the ID performance (\textbf{20.05\%} decay in average). The results also suggest no significant difference in the OOD robustness between generative and discriminative models for text classification. We also provide the results of GPT-3 with in-context learning in Appendix G since it leverages a different training strategy.

\noindent\textbf{Model-level Analysis.} On the model level, we observe that ELECTRA-large achieves the best performance for both ID (\textbf{89.18\%}) and OOD (\textbf{74.62\%}) tests.  Lightweight models, BERT-base, GPT-2, and DistilBERT-base, are in the bottom three on \method with the lowest OOD performance. In contrast, the base-size ELECTRA and ALBERT achieve comparable generalization results. Moreover, by comparing the Friedman rank of OOD and ID tests in Table \ref{tab:overall_per}, we observe that the fluctuation of the OOD F-rank is slightly lower than the ID F-rank, which hints that the uncertainty of performance has been decreased on \method by using a large amount of the test data.

\noindent\textbf{The Performance of Compressed Models.} The results of \method suggest that OOD generalization still faces fundamental challenges, especially for lightweight models. For example, we find that compressed models (e.g., DistilBERT-base) show relatively low performance compared to others. Differently, the OOD performance of ALBERT-base (11M parameters) is significantly higher than DistilBERT-base (\textbf{65.30\% vs. 61.94\%}), even better than several moderate-sized models (BERT-large, GPT2-medium, and XLNet-base). 






\subsection{Discussion}

\noindent\textbf{Human vs. Model.} The average performance decay between in- and out-of-domain tests of humans (87.10\% -- ID vs. 80.14\% -- OOD) is significantly lower than models, even for the best-performing model with the lowest performance decay (\textbf{7.82\% vs. 16.33\%}), as shown in Table \ref{tab:detail}. Regarding the average OOD performance, the human baseline is also much higher than the models, with at least an \textbf{6.69\%} increase (80.14\% vs. 74.62\%)\footnote{Note that the human performance of RTE, MRPC, and QQP is still adjusted and will be updated in the next version.}. Such a large performance gap indicates that PLMs cannot achieve competitive results with humans on \method. More specifically, the human baseline outperforms the state-of-the-art results on five of eight tasks. It is noteworthy that we control OOD evaluations of humans in the same experimental setting with models by testing on unseen samples.

\noindent\textbf{OOD Robustness.} As shown in Table \ref{tab:detail}, we suggest that there is no silver bullet towards the OOD robustness, given that no single model can consistently outperform others over all tasks on \method. For example, ELECTRA-large can only achieve the best performance on four of eight tasks. We also find that the generalization for the CoLA dataset is the most challenging task for models since the test set holds the biggest difference with training data. In contrast, models tend to perform better on the relatively easy dataset, such as sentiment analysis (SST-2). For example, the best-performing model ELECTRA-large can achieve a \textbf{94.67\%} accuracy on SST-2 yet only a \textbf{37.85\%} Matthew's Corr on CoLA. Besides, we also observe that the distribution shift between the ID and OOD datasets largely influences the OOD generalization results. In particular, the performance decay on the OOD test is exacerbated by the increase of distribution shifts, as shown in Appendix \ref{sec:appendixC}.

\noindent\textbf{Model Architectures vs. Parameter Size.} The rightmost column of Table \ref{tab:detail} demonstrates the decreased ratio representing the model robustness to some extent. Although the large-sized model, such as T5-large, and RoBERTa-large, can surpass the corresponding base-sized models in terms of the lower decreased ratio, empirical evidence from Table \ref{tab:detail} also shows that model types could be more influential than the parameter size towards the OOD performance. Specifically, as shown in Table \ref{tab:detail}, results of the same architecture with different parameters are closer to the results of similar parameter-size models based on different architectures. For instance, the decreased ratio of T5 architectures pre-training with different parameter sizes (T5: 16.98\% -- large (737M); 18.47\% -- base (223M); 18.57\% -- small (61M)) are close to each other, similar to RoBERTa (18.46\% --large vs. 19.40\% -- base). It hints that designing model architectures and training methods could be one of the future directions for improving OOD robustness.


\begin{table}[t]
\center
\small
\begin{tabular}{lccc}
\toprule
\textbf{Model} & \textbf{Precision} & \textbf{Recall} & \textbf{F1} \\ \midrule
ELECTRA-small   & \textbf{15.17}  & 42.48  & \textbf{20.54} \\
ELECTRA-large   & 14.61     &\textbf{43.96}  & 20.23 \\
BERT-large      & 12.97     & 40.33  & 18.14 \\
ALBERT-base     & 13.12     & 39.43  & 18.10 \\
T5-large        & 12.78     & 39.09  & 17.74 \\
ELECTRA-base    & 12.87     & 37.73  & 17.66 \\
T5-base         & 12.69     & 37.98  & 17.42 \\
BART-base       & 12.16     & 36.76  & 16.79 \\
BERT-base       & 11.97     & 36.17  & 16.53 \\
T5-small        & 12.03     & 35.41  & 16.49 \\
BART-large      & 11.68     & 36.23  & 16.25 \\
XLNet-large     & 11.63     & 36.38  & 16.18 \\
RoBERTa-base    & 10.68     & 34.29  & 14.94 \\
DistilBERT-base & 10.56     & 31.87  & 14.55 \\
GPT2-large      & 10.24     & 31.54  & 14.10  \\
GPT2-medium     & 10.20     & 30.68  & 13.94 \\
XLNet-base      & 9.86      & 30.90   & 13.68 \\
GPT2            & 9.63      & 28.53  & 13.11 \\
RoBERTa-large   & 7.93      & 27.88  & 11.45 \\ \bottomrule
\end{tabular}
\caption{The average F1 score of the rationale overlap on three sentiment analysis tasks sorts the table.}
\label{tab:decay_ratio}
\end{table}

\begin{figure*}[ht]
\centering
\subfigure[COLA]{
\begin{minipage}[t]{0.32\linewidth}
\centering
\includegraphics[width=\textwidth]{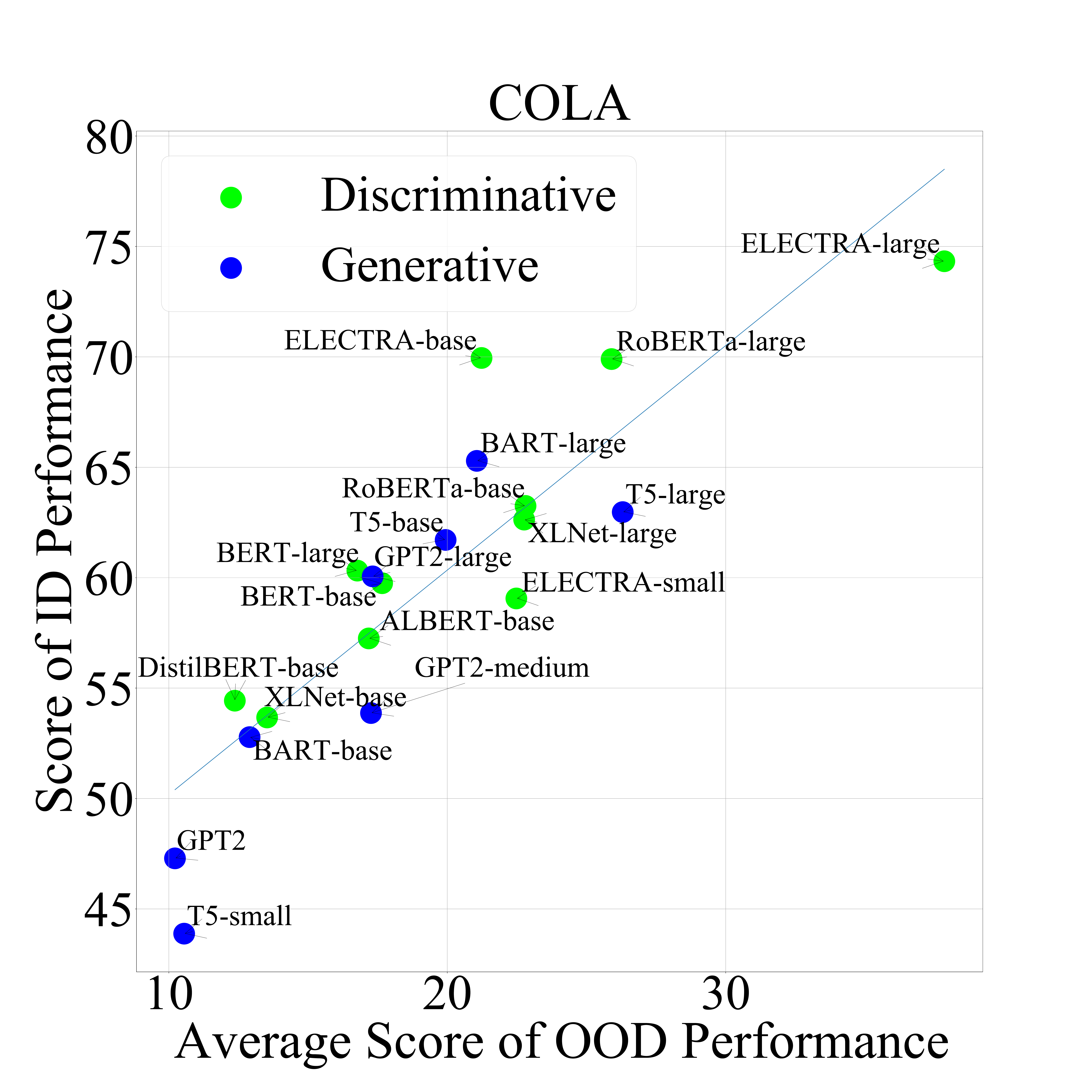}
\end{minipage}%
}%
\subfigure[MNLI]{
\begin{minipage}[t]{0.32\linewidth}
\centering
\includegraphics[width=\textwidth]{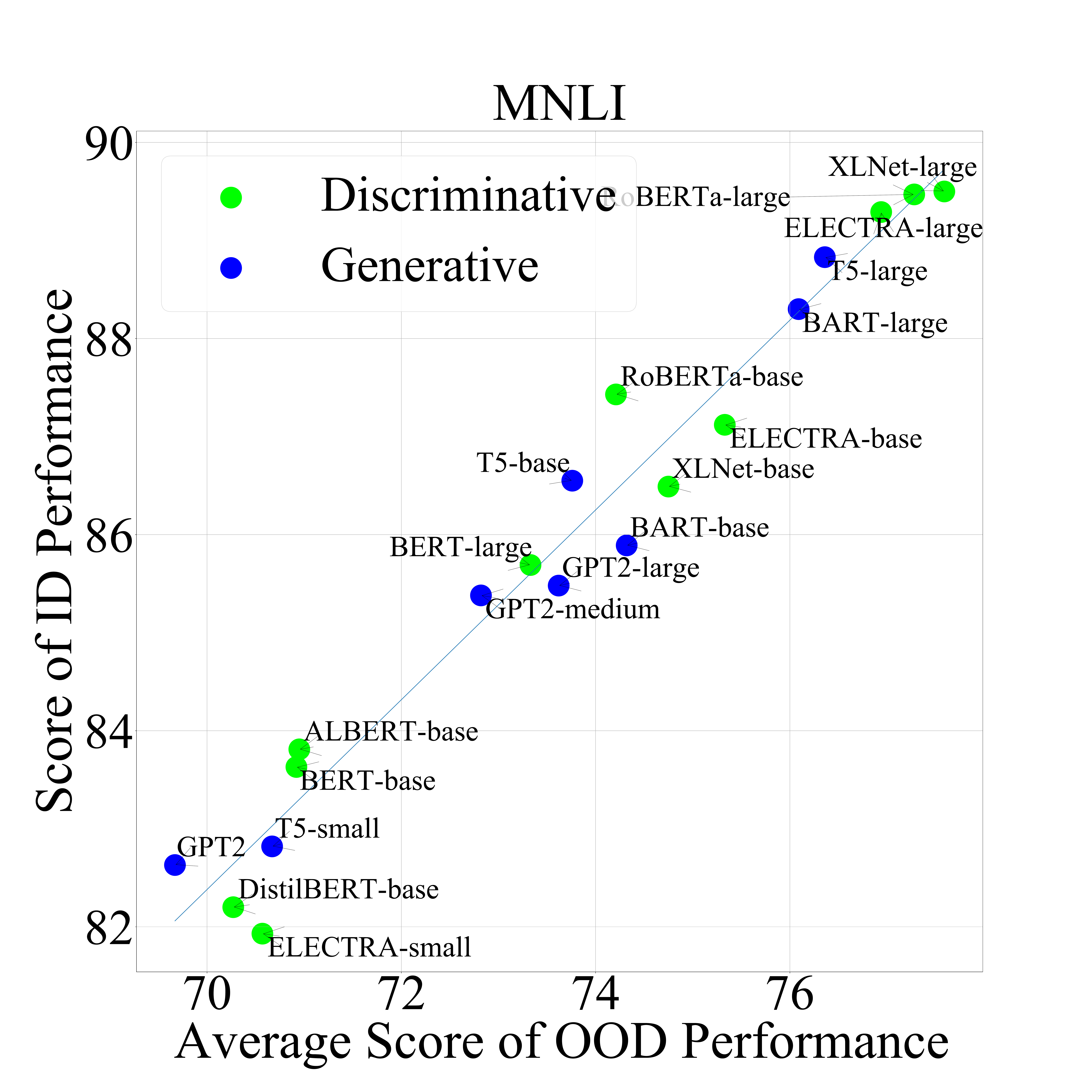}
\end{minipage}%
}
\subfigure[MRPC]{
\begin{minipage}[t]{0.32\linewidth}
\centering
\includegraphics[width=\textwidth]{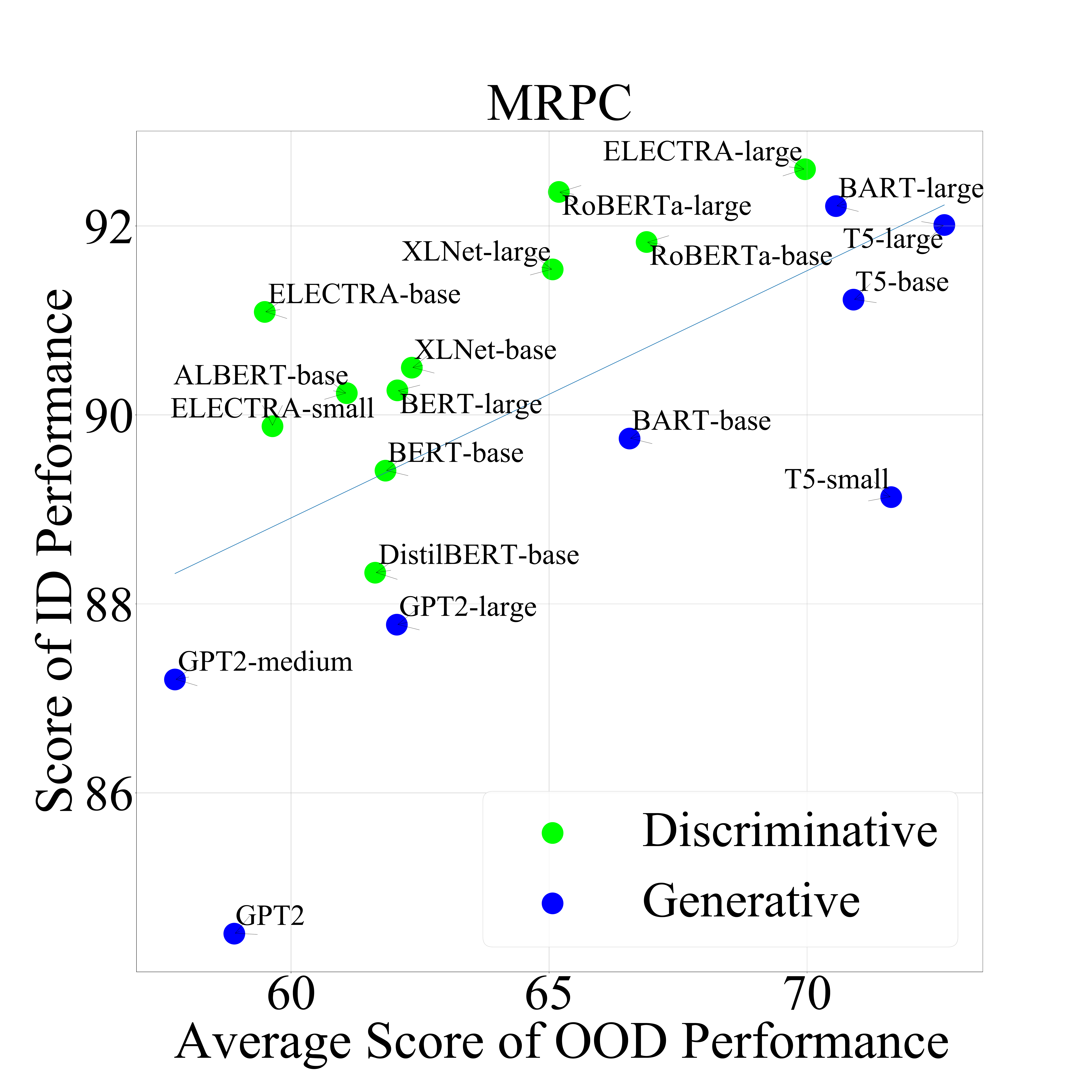}
\end{minipage}%
}
\caption{Scatter figures that illustrate the correlation between ID and OOD performance for different tasks.}
\label{fig:scatter}
\end{figure*}

\begin{figure}[t]
\centering 
\small
{%
\includegraphics[width=.36\textwidth]{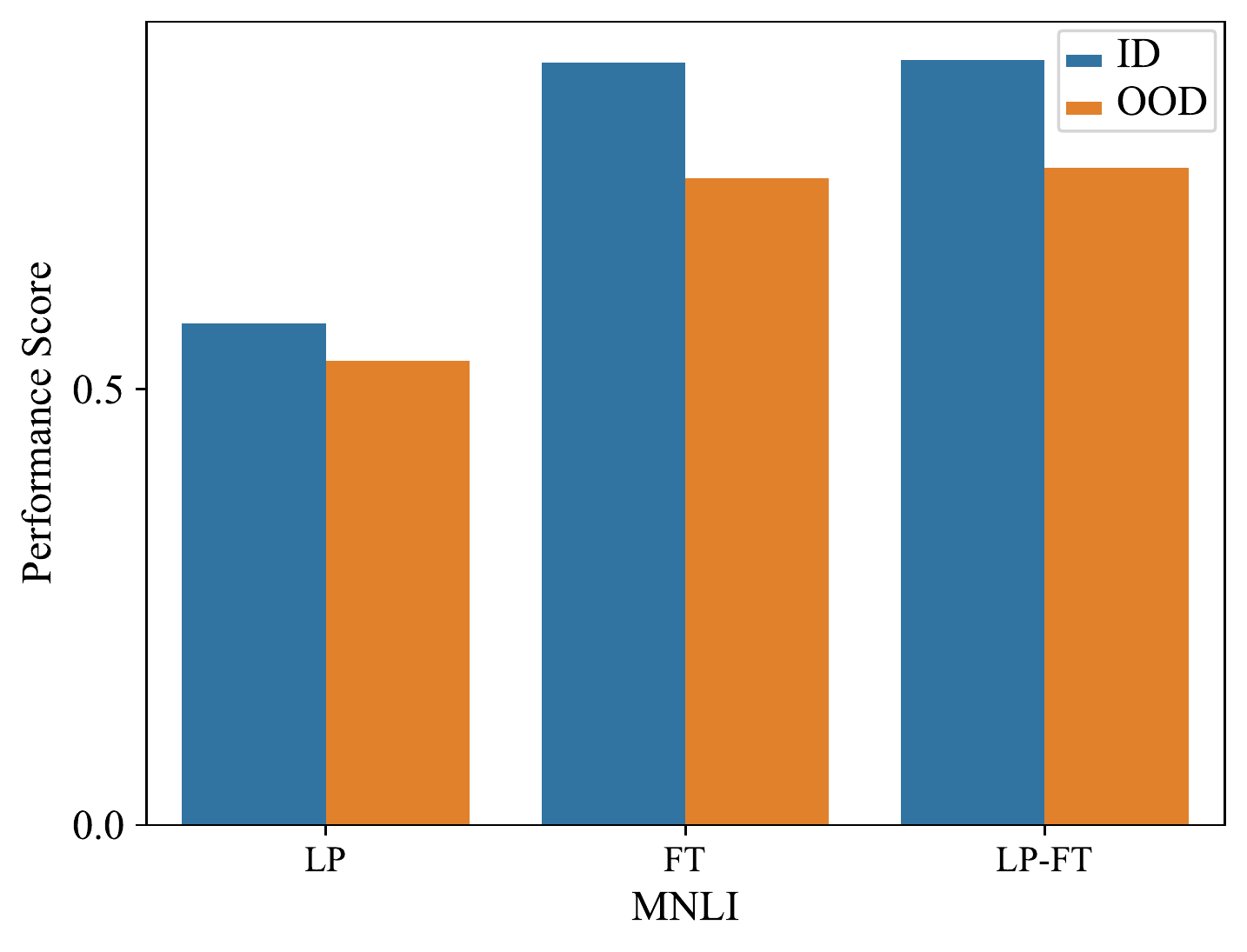} 
}
\caption{The influence of different tuning strategies on the task of MNLI, including Linear probing, Fine-tuning, and Linear probing then Fine-tuning (LP-FT). The results are based on RoBERTa-base.} 
\label{fig:tuning}
\end{figure}

\noindent\textbf{Robustness vs. Trust.} The average results of the rationale overlap between models and humans for three sentiment analysis tasks are shown in Table \ref{tab:decay_ratio}, indicating the trust measurement. As shown in the table, somehow surprisingly, we find that the best-performing discriminative model -- RoBERTa-large (see Table \ref{tab:overall_per}) achieves the lowest rationale overlap between humans and models. While RoBERTa-large can achieve a relatively high overlap with humans on the NLI task (see Appendix D). This can be because the rationale overlap is largely influenced by datasets.

It is noteworthy that small-sized models can achieve relatively higher rationale overlaps than large-sized models, which is generally consistent with the results reported by previous work \cite{deyoung2020eraser}. For instance, ELECTRA-small achieves the highest F1 score with only 13.48M parameters. In addition, the models pre-trained with the same architectures usually achieve similar performance (e.g., ELECTRA-small and ELECTRA-large, GPT2-medium and GPT2-large).

\noindent\textbf{ID vs. OOD Performance.} We show the correlation of three tasks between the in- and out-of-domain results in Figure \ref{fig:scatter} (the full results can be found at Appendix \ref{sec:appendixD}). Unsurprisingly, we observe that the in-domain performance is always higher than the out-of-domain performance. Specifically, we find that the OOD performance is much lower than the ID performance in the task of COLA. In contrast, the gap between ID and OOD performance based on SST-2 and MNLI is relatively lower than others. We suppose this is partially influenced by the distribution shift between the in- and out-of-domain datasets. 

Regarding the type of pre-trained models, we show that discriminative models show a stronger linear correlation when compared to generative models (19 data points). From the task perspective, we observe that datasets largely influence the correlation between ID and OOD. For instance, ID and OOD performance are inversely correlated on MRPC yet almost correlated on other tasks, hinting that the inverse correlation is possible for the specific task when the size of test samples is limited. 

\noindent\textbf{The Influence of Tuning Methods.} Taking MNLI as an example, we compare the results of RoBERTa-base using three different training strategies in Figure \ref{fig:tuning}. As found by previous work \cite{kumar2022finetuning}, fine-tuning can do worse than linear probing in the presence of a large distribution shift in CV. However, as shown in Figure \ref{fig:tuning}, we find that linear probing methods show relatively low accuracy for both ID and OOD tests, which is different from the conclusion in CV. This can be because freezing pre-trained features hinders the generalization of NLP tasks that are generally more complex than the OOD generalization in CV. While the LP-FT can be relatively helpful for improving the OOD robustness of NLP models in terms of the slight performance improvement compared to the standard fine-tuning method. For this reason, there is still much room to improve in designing methodologies of domain generalization that can improve the OOD robustness for text classification. In addition to tuning methods discussed in \method, the recently emerging trend of the development of large-scale language models (LLMs) represented by ChatGPT is worth paying attention to. In particular, how to appropriately define the OOD generalization for LLMs is still under-explored since the pre-training corpus of these models is not disclosed yet \cite{wang2023robustness}.

\section{Conclusion}

We constructed \method, an OOD robustness benchmark for natural language understanding tasks that aim to enable fair evaluation over multiple datasets from multiple domains in a consistent setting. With \method, we evaluate 21 pre-trained models on 8 classification tasks, providing analysis using 3 different tuning strategies and post-hoc analysis for gaining internal causes for the OOD robustness. We conclude that (1) current PLMs still have a lag much behind human-level towards the OOD robustness; (2) the ID and OOD performance usually hold a linear correlation in most cases, while the coefficiency of the correlation is primarily related to the selection of OOD datasets; (3) stronger architectures can bring decent performance benefit, especially for the OOD performance. 


\section*{Limitation}

Our primary focus is on the OOD robustness of text classification tasks. However, there are other NLP tasks that the community should not ignore. \method currently does not include language generation tasks such as machine translation, summarization, and dialogue. Moreover, extending the current \method to more real-world datasets from different domains is of great importance. We aim to make \method a continuously maintained project.

\section*{Ethics Statement}

This paper honors the ACL Code of Ethics. Public available datasets are used to establish the GLUE-X leaderboard. No private data was used. All annotators from the crowdsourcing company have received enough labor fees corresponding to their amount of annotated instances. The code and data are open-sourced under the CC-BY-NC-SA license.

\section*{Acknowledgement}
We acknowledge with thanks Wei Zhou from Zhejiang University, who help us build the website, as well as the many others who have helped. We would also like to thank anonymous reviewers for their insightful comments and suggestions to help improve the paper, especially for Reviewer 2. This publication has emanated from research conducted with the financial support of the Pioneer and ``Leading Goose'' R\&D Program of Zhejiang under Grant Number 2022SDXHDX0003 and the 72nd round of the Chinese Post-doctoral Science Foundation project 2022M722836. Yue Zhang is the corresponding author.

\bibliography{custom}
\bibliographystyle{acl_natbib}

\newpage
\appendix





\section{Data Collection} \label{sec:appendixA}
We derive the CoLA-OOD dataset from the Public High School English Exam, which contains 304,277 examples. The original multi-choice fill-in tests are converted into COLA-style, with correct answers as positive examples and incorrect answers as negative examples. The golden answer is given by the English teacher who is a native speaker or holds an English Teaching degree. We collect the data from publicly available internet resources, and the original open-access materials can be found from \url{https://www.koolearn.com/shiti}.

The input of the CoLA-OOD dataset, Grammar Test, is a text span containing a QA pair or a few sentences. The ground truth of the output is to decide whether the grammar of the sentence is acceptable or not. For example, given the sentence `Is there a post office near here?  Yes, there isn't .', the label is unacceptable since there is a grammar error existing in the input. Otherwise, for a sentence without any grammar errors, `The young man is the CEO of the company, In other words, he is in charge of the company.', the corresponding label is acceptable. 

\section{Training Details} \label{sec:appendixB}

\begin{figure}[htbp]
\centering 
{%
\includegraphics[width=.45\textwidth]{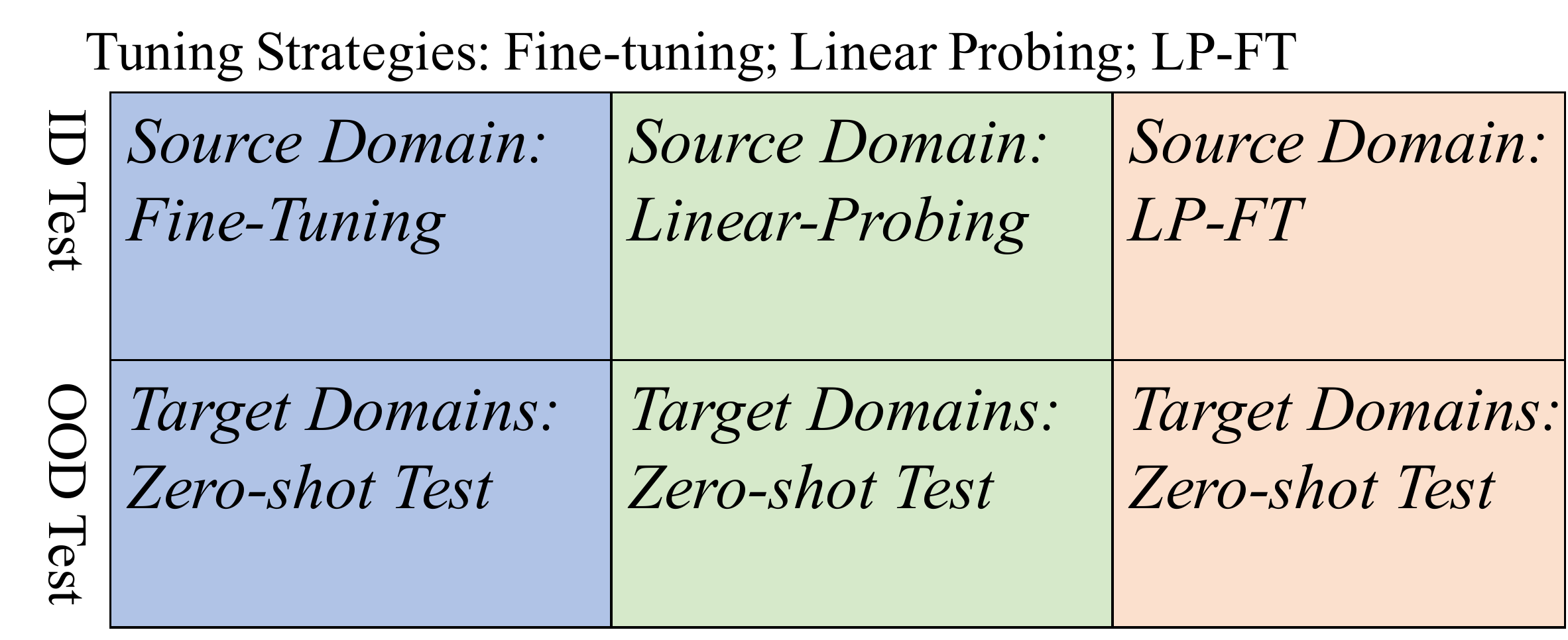} 
}
\caption{The demonstration of training and testing settings used for cross-domain evaluations in \method.} 
\label{fig:settings}
\end{figure}

We illustrate the cross-domain evaluation settings used for \method in Figure \ref{fig:settings}. Notably, the source domain only contains a single dataset, while target domains can include more than one dataset from multiple domains.

Regarding the training, we performed the grid search for each task, kept the best-performing checkpoint in ID datasets, and tested their performance on their corresponding OOD datasets. The hyperparameters used by these weights can be seen in Table \ref{tab:Hyperparameters}.

\section{Domain Distributions} \label{sec:appendixC}
We evaluate distribution shifts between different datasets regarding Maximum Mean Discrepancy(MMD) and Word Overlap Rate. MMD distance focuses on the semantic distribution shift between datasets, while Word Overlap Rate pays more attention to superficial similarity.
\begin{table}[t]
\centering
\small
\begin{tabular}{lccc}
\toprule
\textbf{Model}  & \textbf{F1} & \textbf{Precision} & \textbf{Recall}\\ \midrule
ELECTRA-base    & \textbf{34.98} & \textbf{31.06}     & 52.94  \\
RoBERTa-large   & 34.89 & 30.57     & \textbf{53.95}  \\
XLNet-large     & 34.73 & 30.64     & 53.32  \\
ELECTRA-large   & 34.37 & 30.34     & 52.67  \\
RoBERTa-base    & 33.78 & 30.06     & 51.36  \\
T5-large        & 33.70 & 29.41     & 52.00  \\
GPT2-large      & 33.37 & 29.95     & 49.51  \\
XLNet-base      & 33.16 & 29.42     & 49.95  \\
GPT2-medium     & 33.06 & 29.46     & 49.65  \\
BERT-large      & 32.96 & 29.45     & 49.88  \\
DistilBERT-base & 32.71 & 29.40     & 48.86  \\
GPT2            & 32.36 & 29.10     & 48.29  \\
ALBERT-base     & 32.34 & 28.93     & 49.09  \\
BART-base       & 32.31 & 28.90     & 49.51  \\
T5-base         & 32.29 & 28.50     & 49.30  \\
ELECTRA-small   & 31.96 & 28.68     & 47.83  \\
BERT-base       & 31.52 & 28.30     & 47.26  \\
BART-large      & 31.31 & 28.04     & 47.80  \\
T5-small        & 30.93 & 27.45     & 47.30  \\
\bottomrule
\end{tabular}
\caption{The rationale overlap based on e-SNLI sorted by descending order of the F1 score.}
\label{tab:rationale_snli}
\end{table}

\begin{figure*}[t]
\centering

\subfigure[COLA]{
\begin{minipage}[t]{0.25\linewidth}
\centering
\includegraphics[width=.9\textwidth]{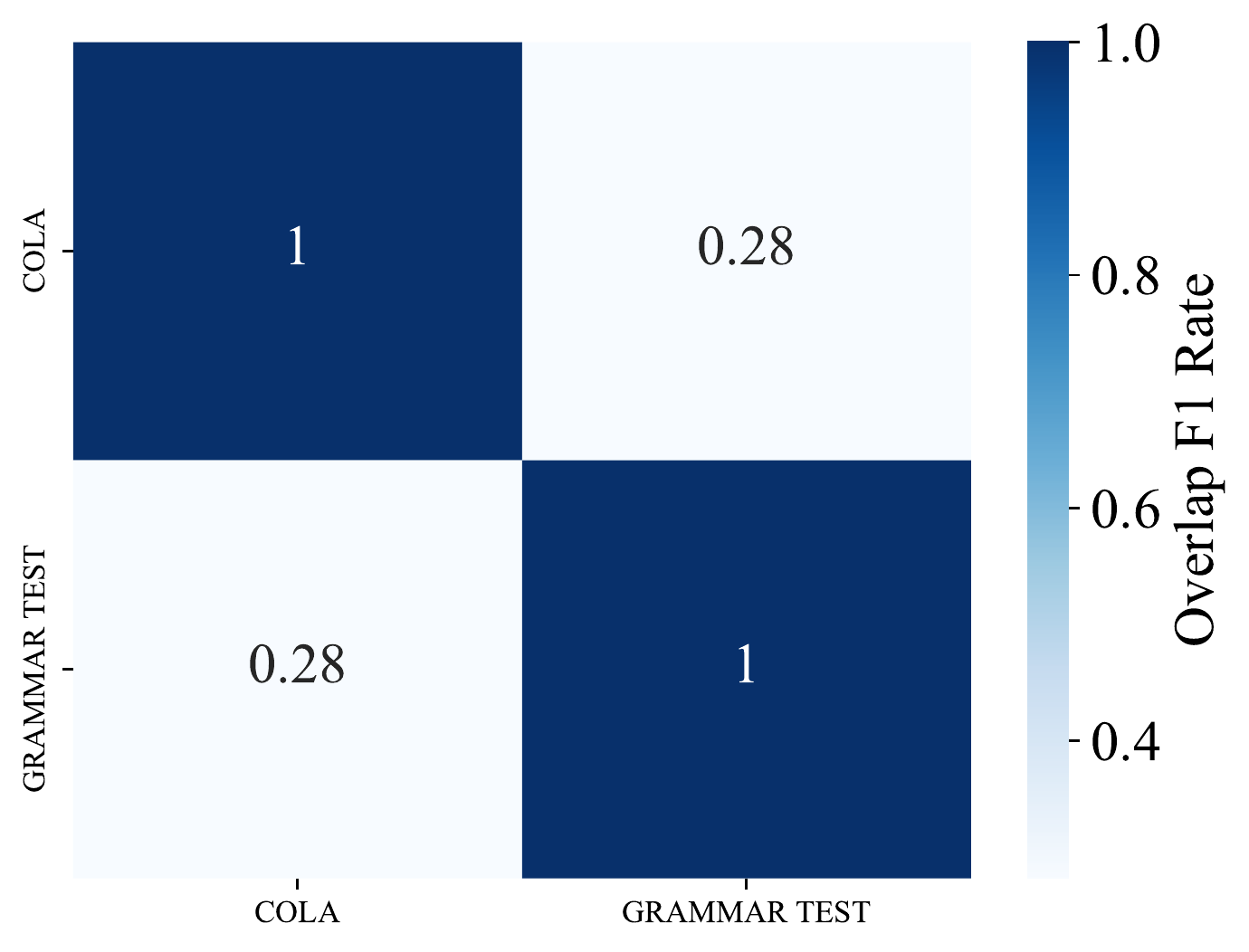}
\end{minipage}%
}%
\subfigure[MNLI]{
\begin{minipage}[t]{0.25\linewidth}
\centering
\includegraphics[width=.9\textwidth]{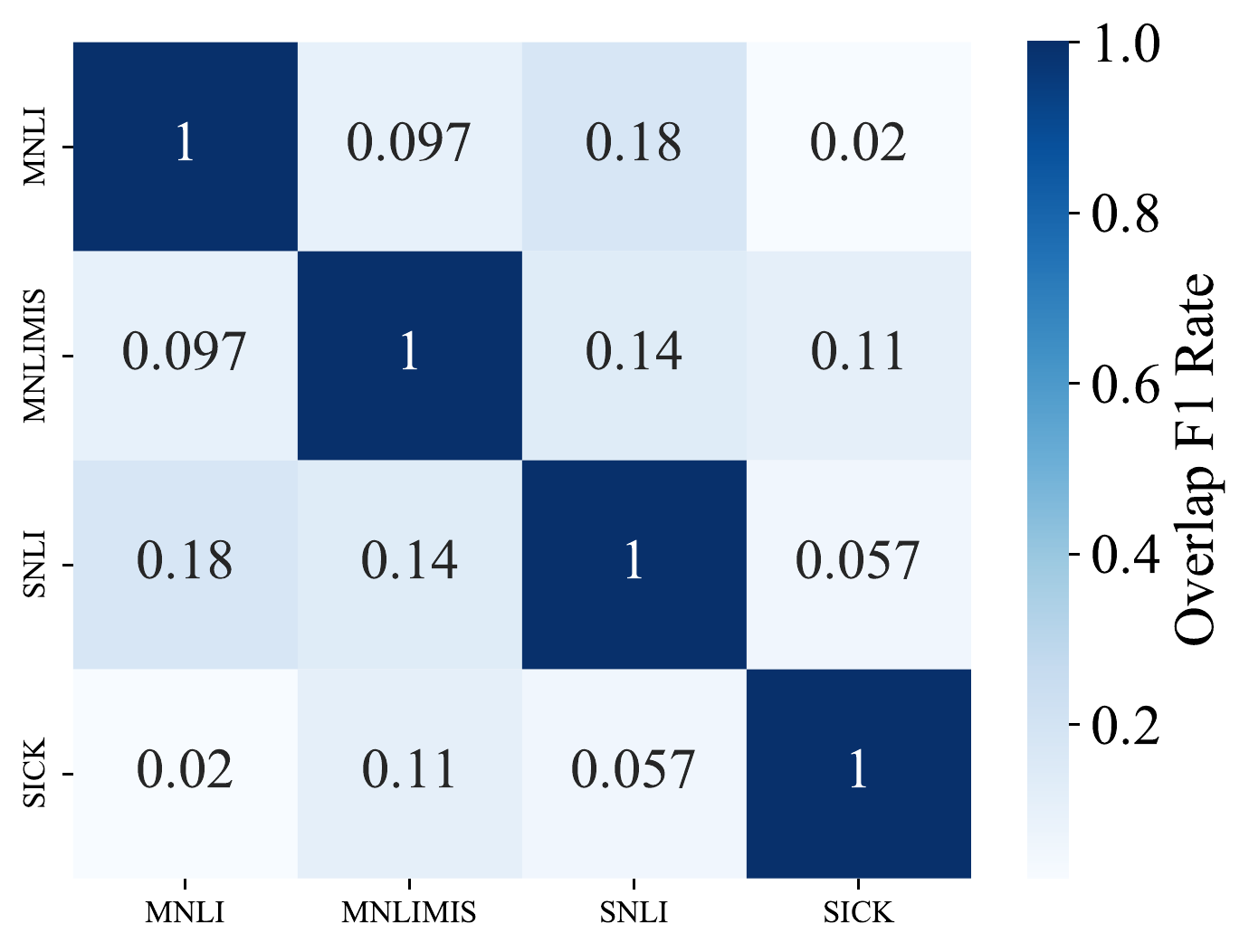}
\end{minipage}%
}%
\subfigure[MRPC]{
\begin{minipage}[t]{0.25\linewidth}
\centering
\includegraphics[width=.9\textwidth]{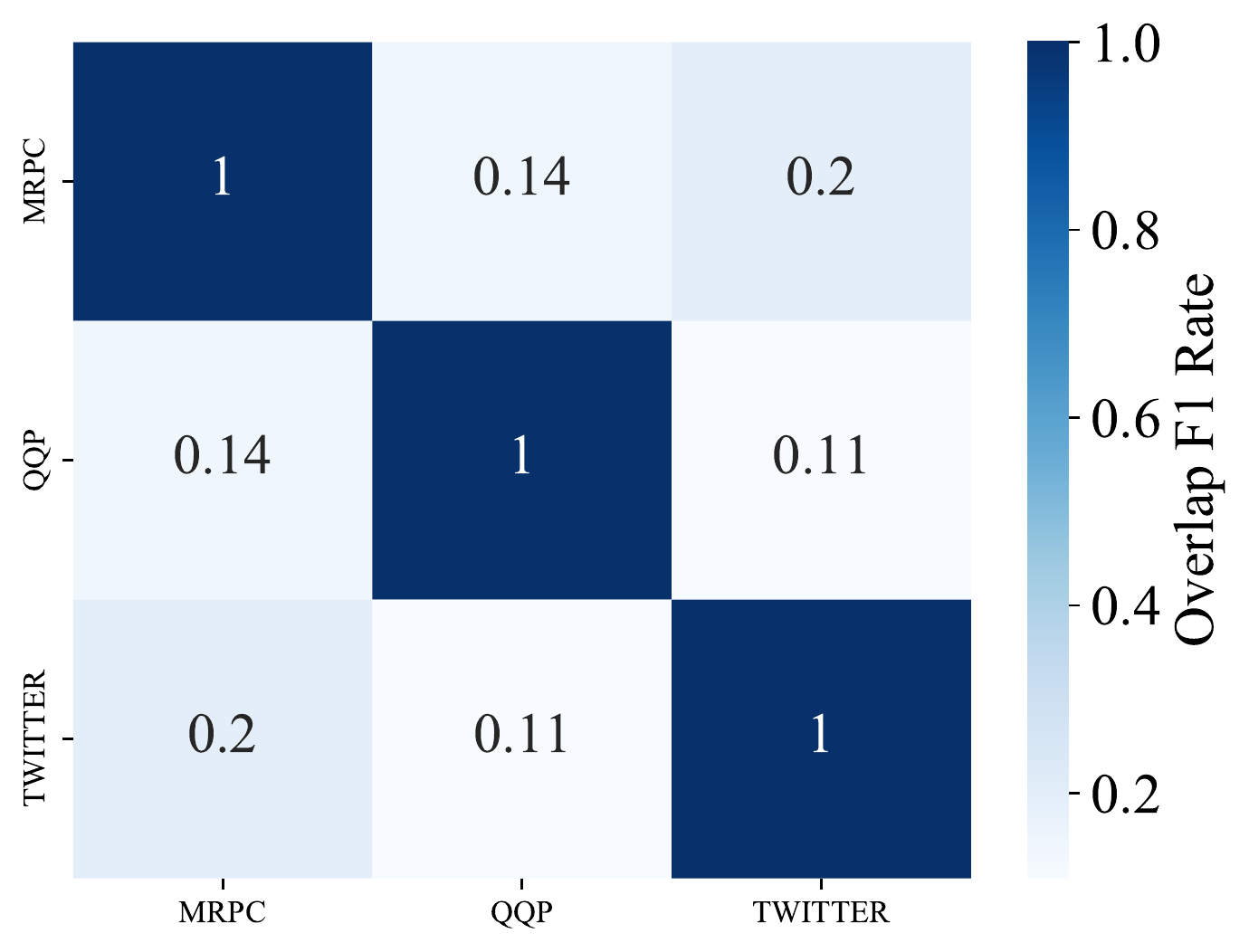}
\end{minipage}%
}%
\subfigure[QQP]{
\begin{minipage}[t]{0.25\linewidth}
\centering
\includegraphics[width=.9\textwidth]{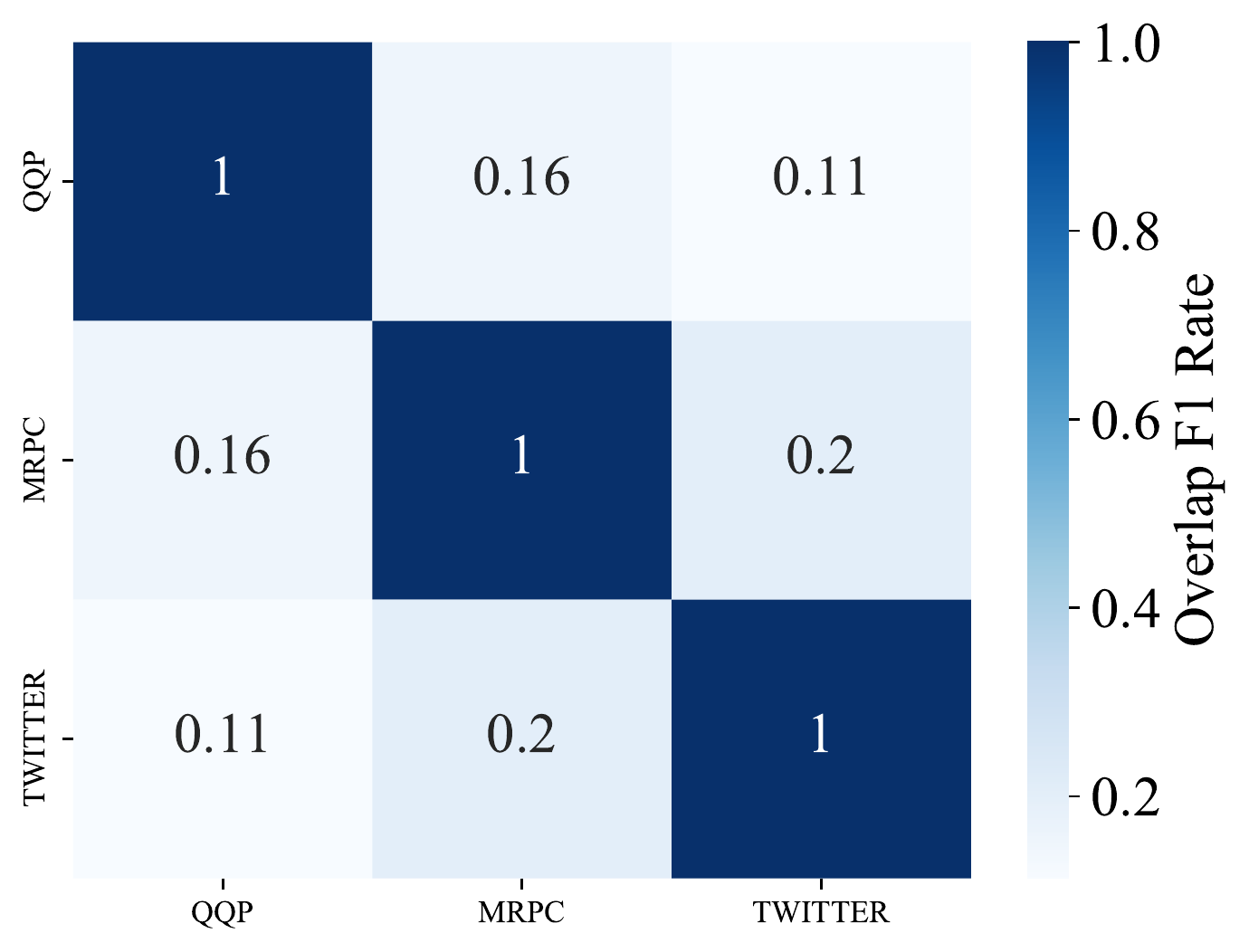}
\end{minipage}%
}%
                 
\subfigure[RTE]{
\begin{minipage}[t]{0.25\linewidth}
\centering
\includegraphics[width=.9\textwidth]{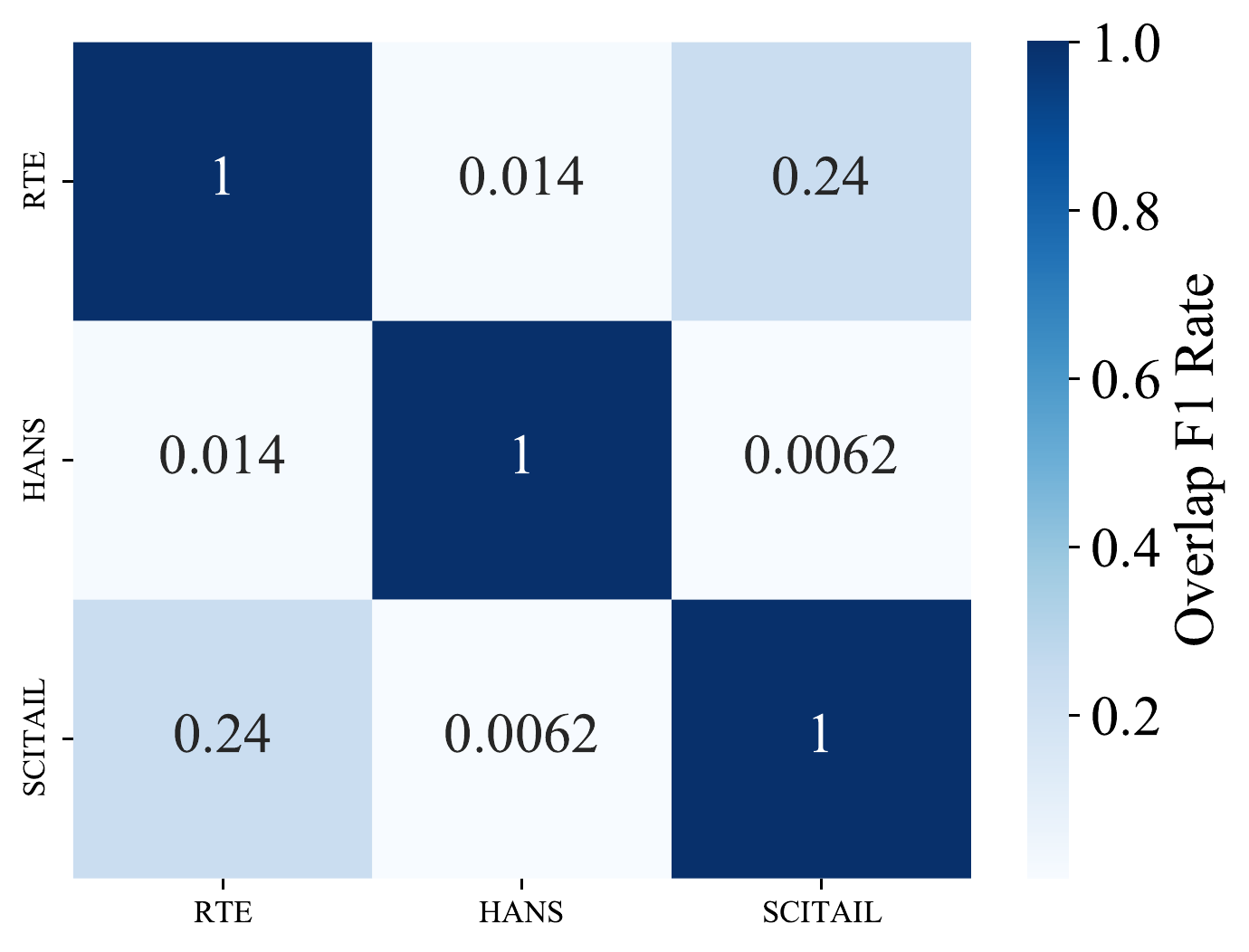}
\end{minipage}
}%
\subfigure[SST-2]{
\begin{minipage}[t]{0.25\linewidth}
\centering
\includegraphics[width=.9\textwidth]{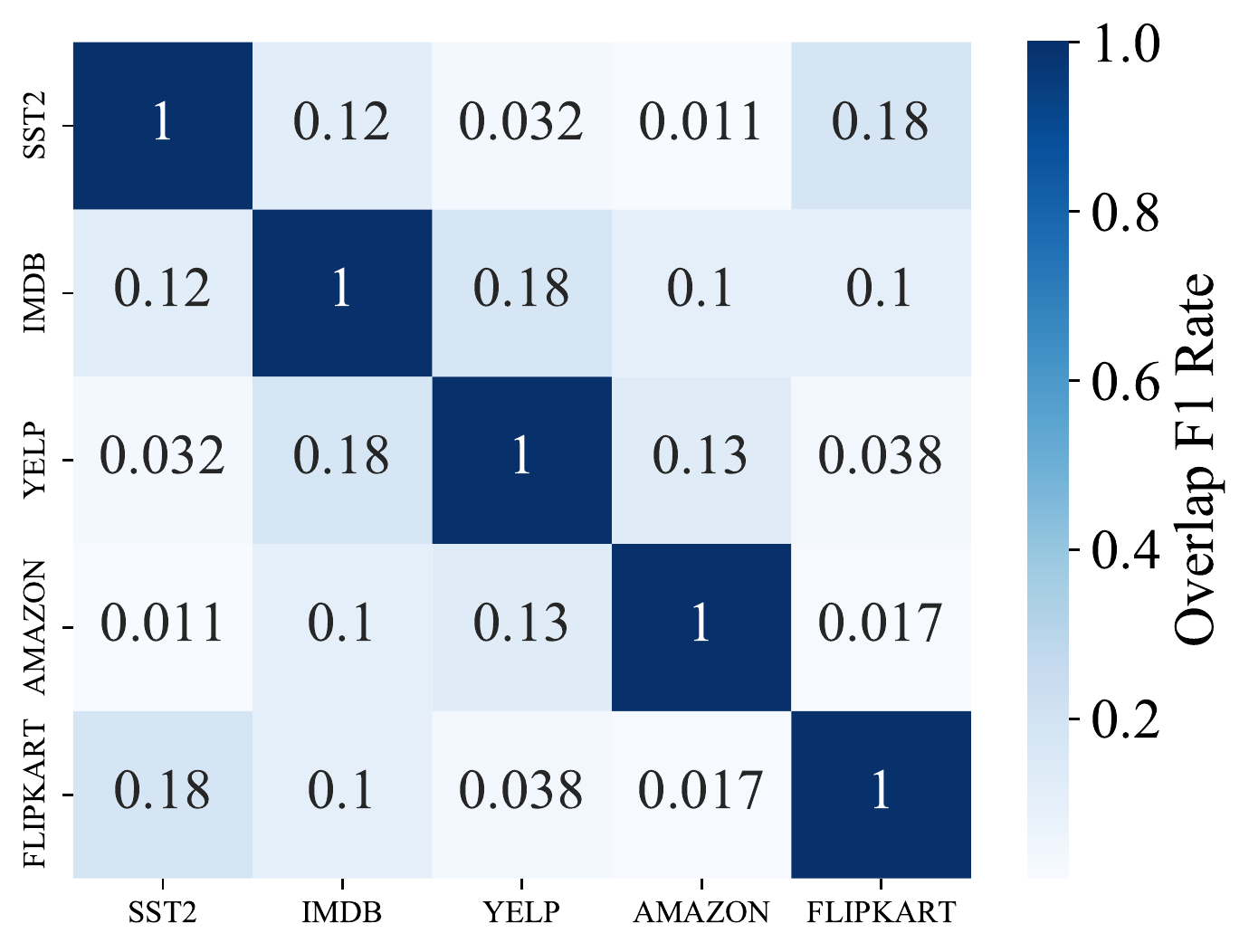}
\end{minipage}
}%
\subfigure[STSB]{
\begin{minipage}[t]{0.25\linewidth}
\centering
\includegraphics[width=.9\textwidth]{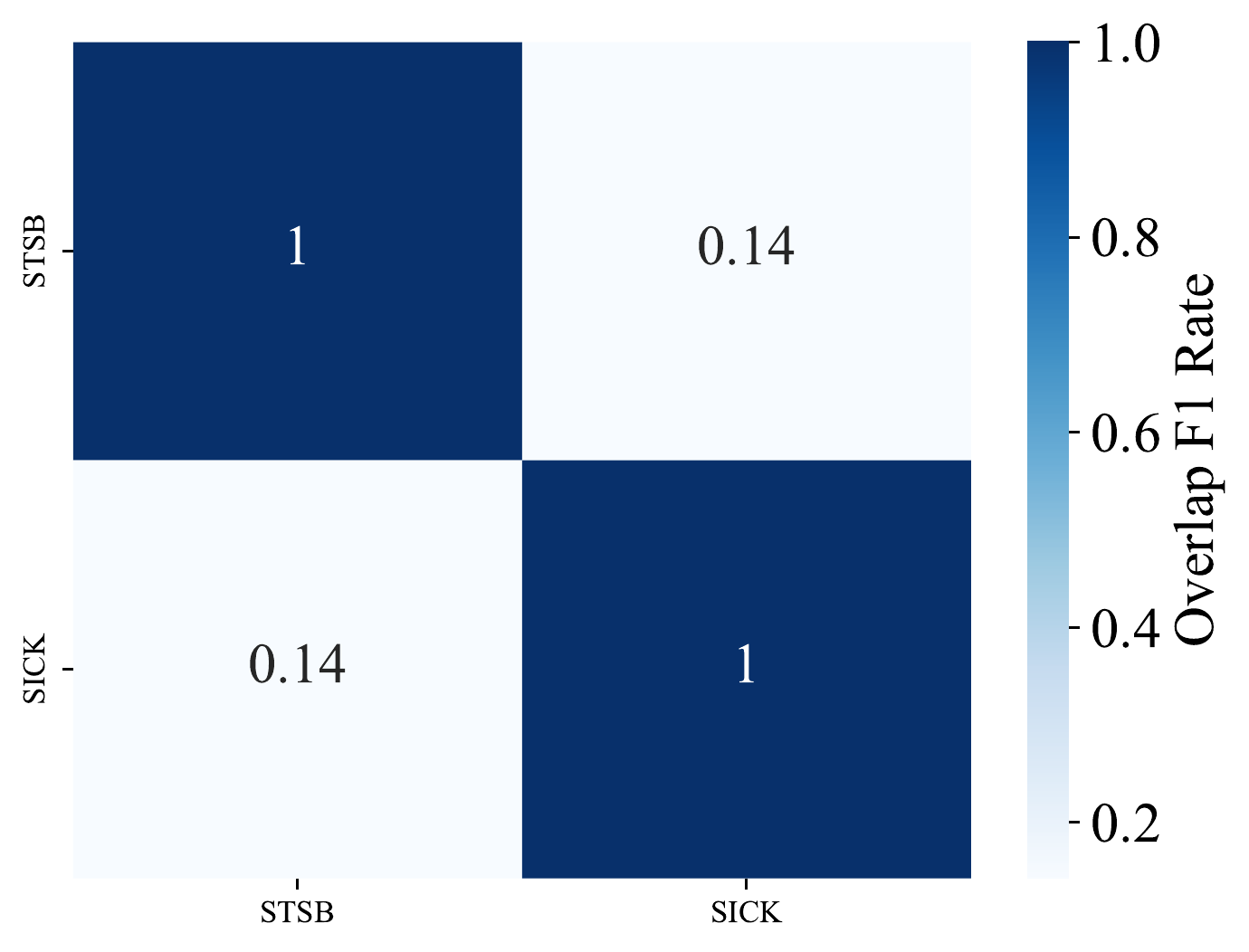}
\end{minipage}
}%
\subfigure[QNLI]{
\begin{minipage}[t]{0.25\linewidth}
\centering
\includegraphics[width=.9\textwidth]{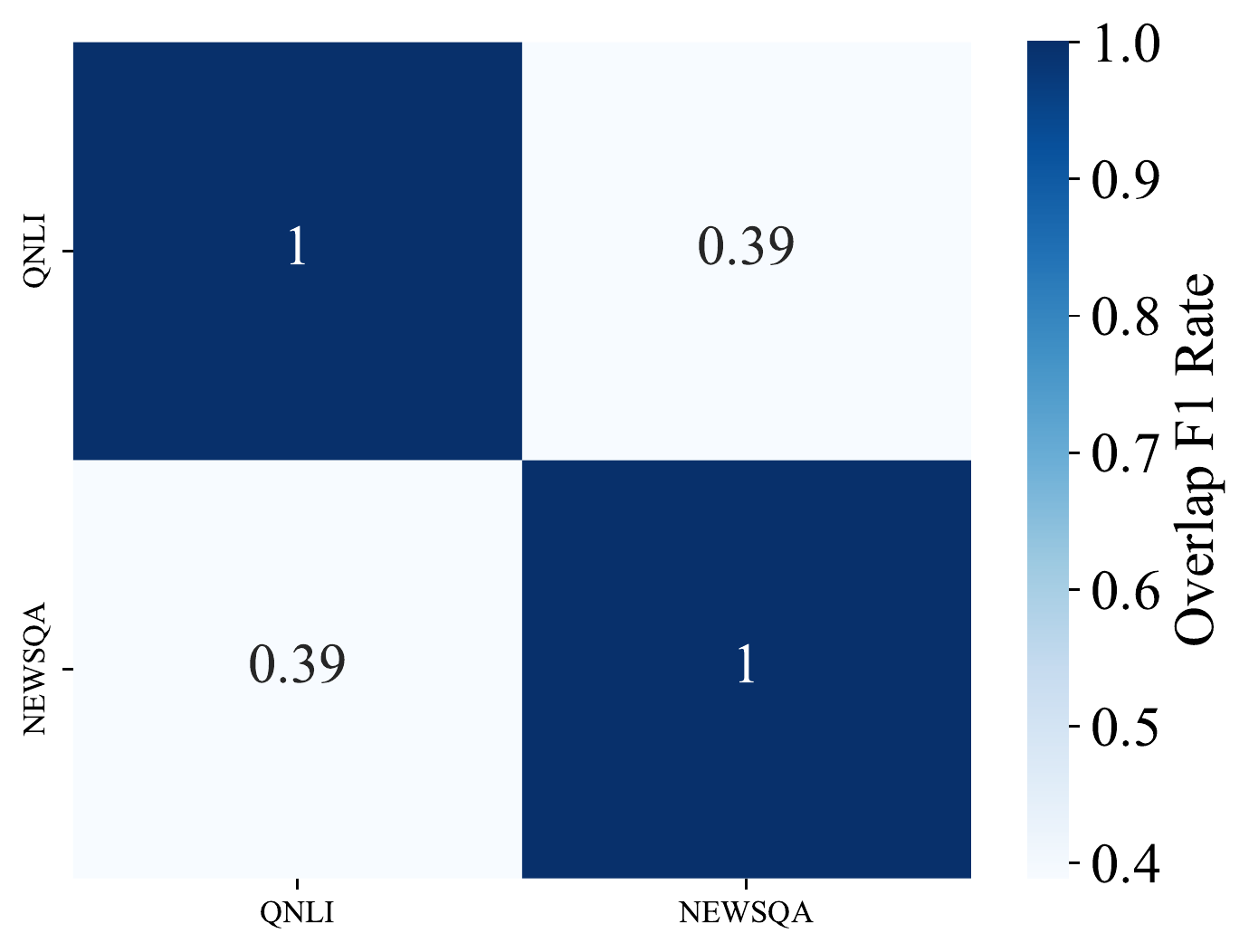}
\end{minipage}
}%
\centering
\caption{The word-level overlap between the training set and test set for each task.}
\label{fig:Word_Overlap}
\end{figure*}

\begin{figure*}[t]
\centering

\subfigure[COLA]{
\begin{minipage}[t]{0.25\linewidth}
\centering
\includegraphics[width=.9\textwidth]{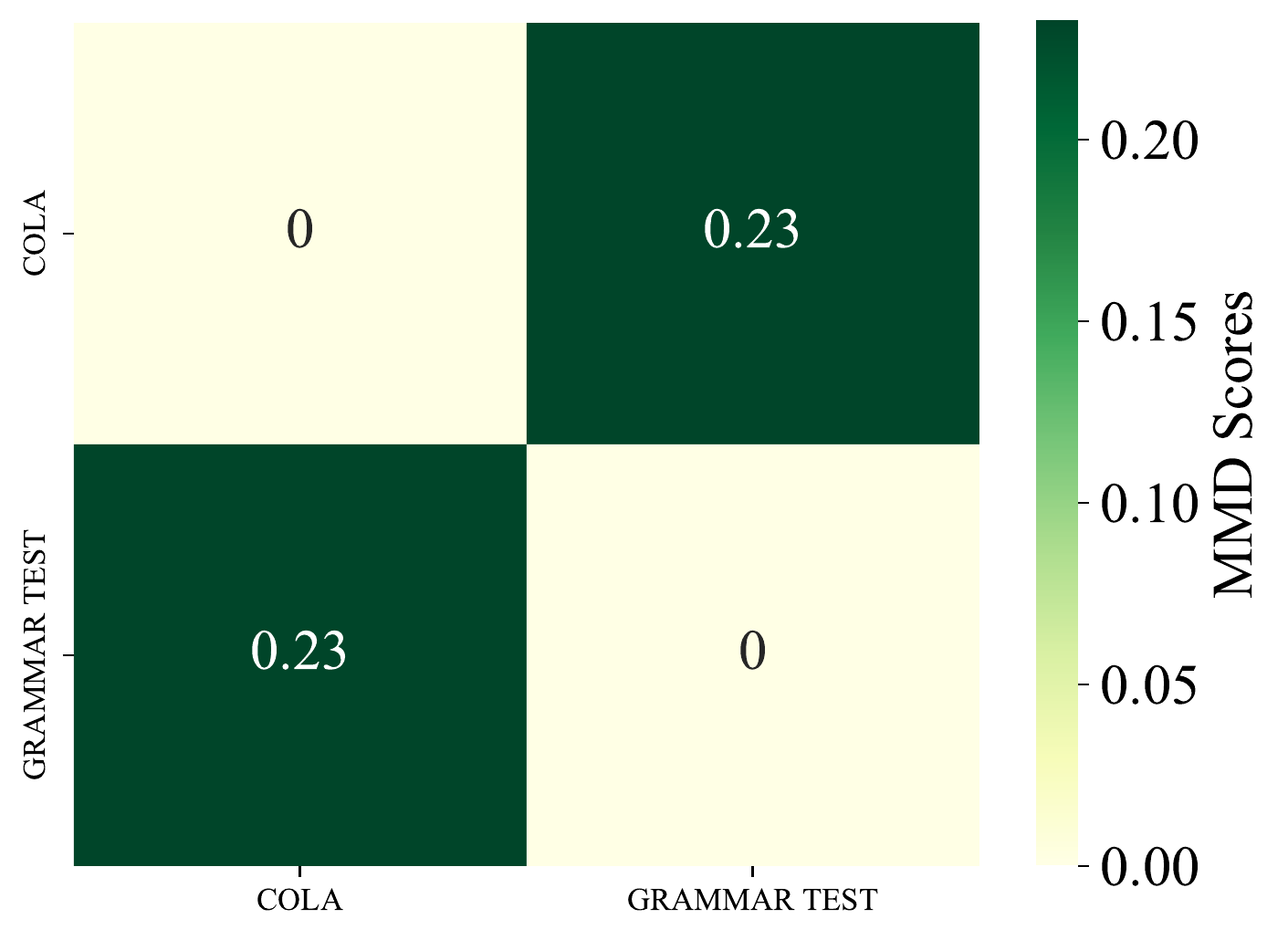}
\end{minipage}%
}%
\subfigure[MNLI]{
\begin{minipage}[t]{0.25\linewidth}
\centering
\includegraphics[width=.9\textwidth]{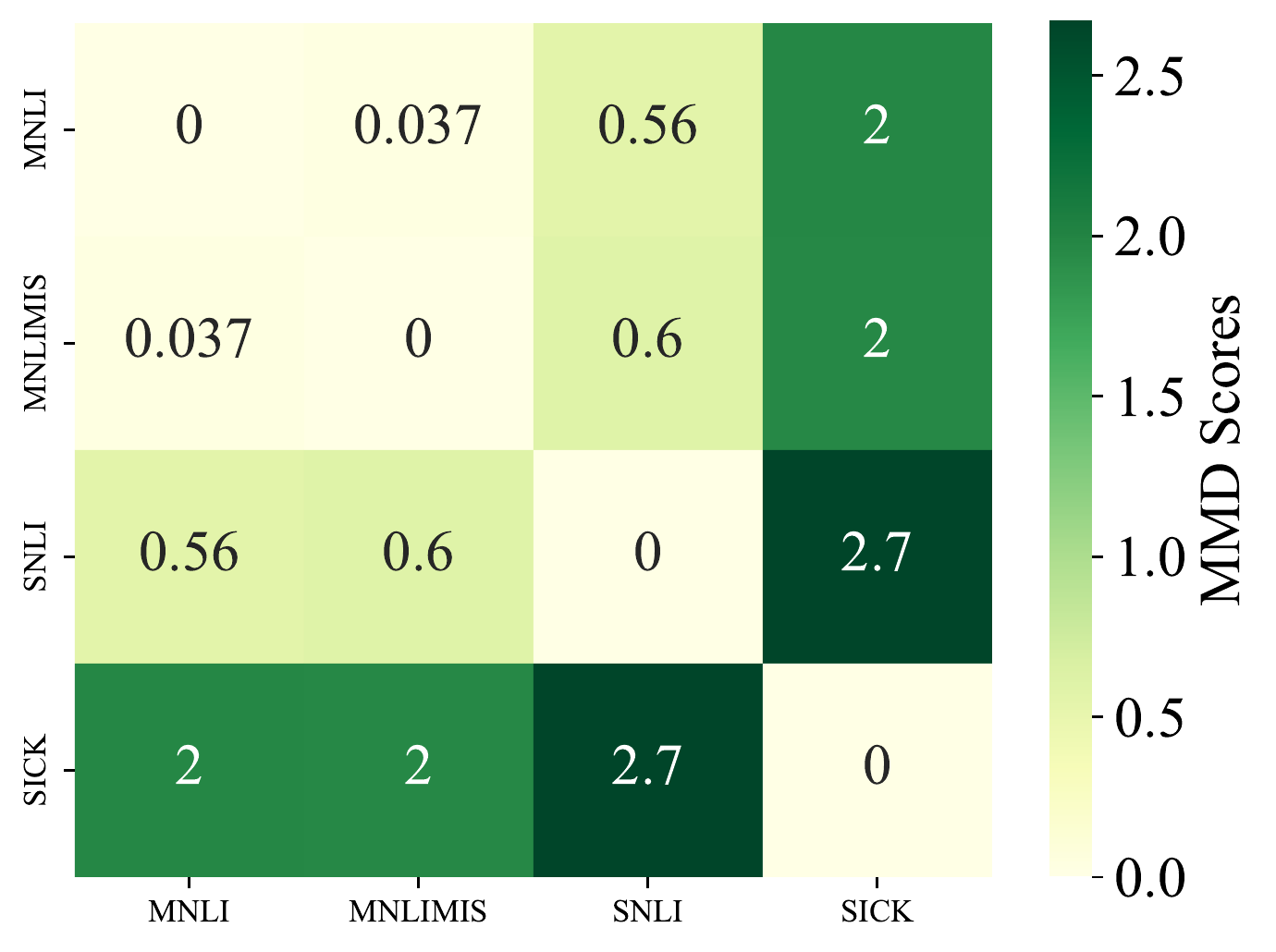}
\end{minipage}%
}%
\subfigure[MRPC]{
\begin{minipage}[t]{0.25\linewidth}
\centering
\includegraphics[width=.9\textwidth]{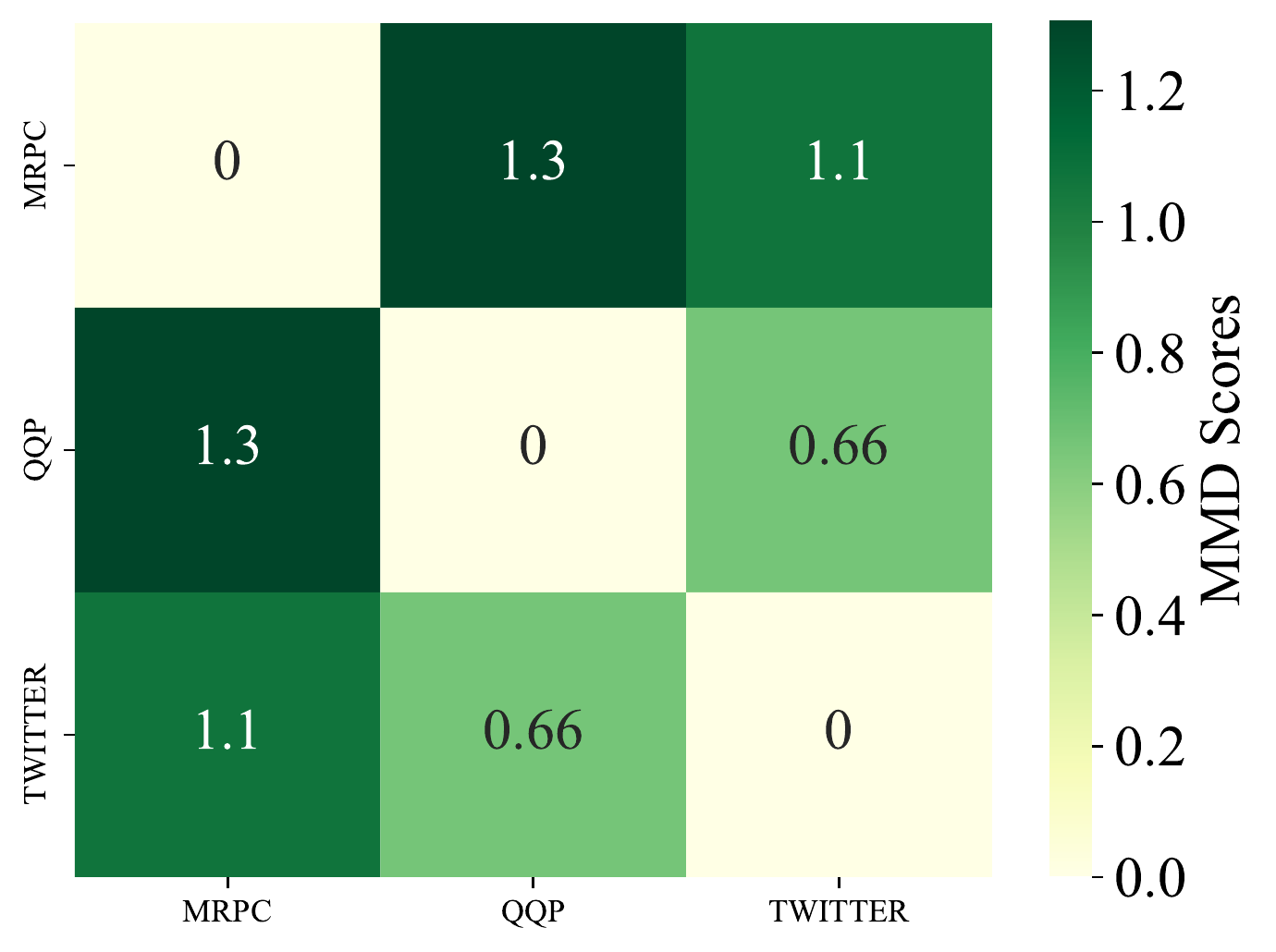}
\end{minipage}%
}%
\subfigure[QQP]{
\begin{minipage}[t]{0.25\linewidth}
\centering
\includegraphics[width=.9\textwidth]{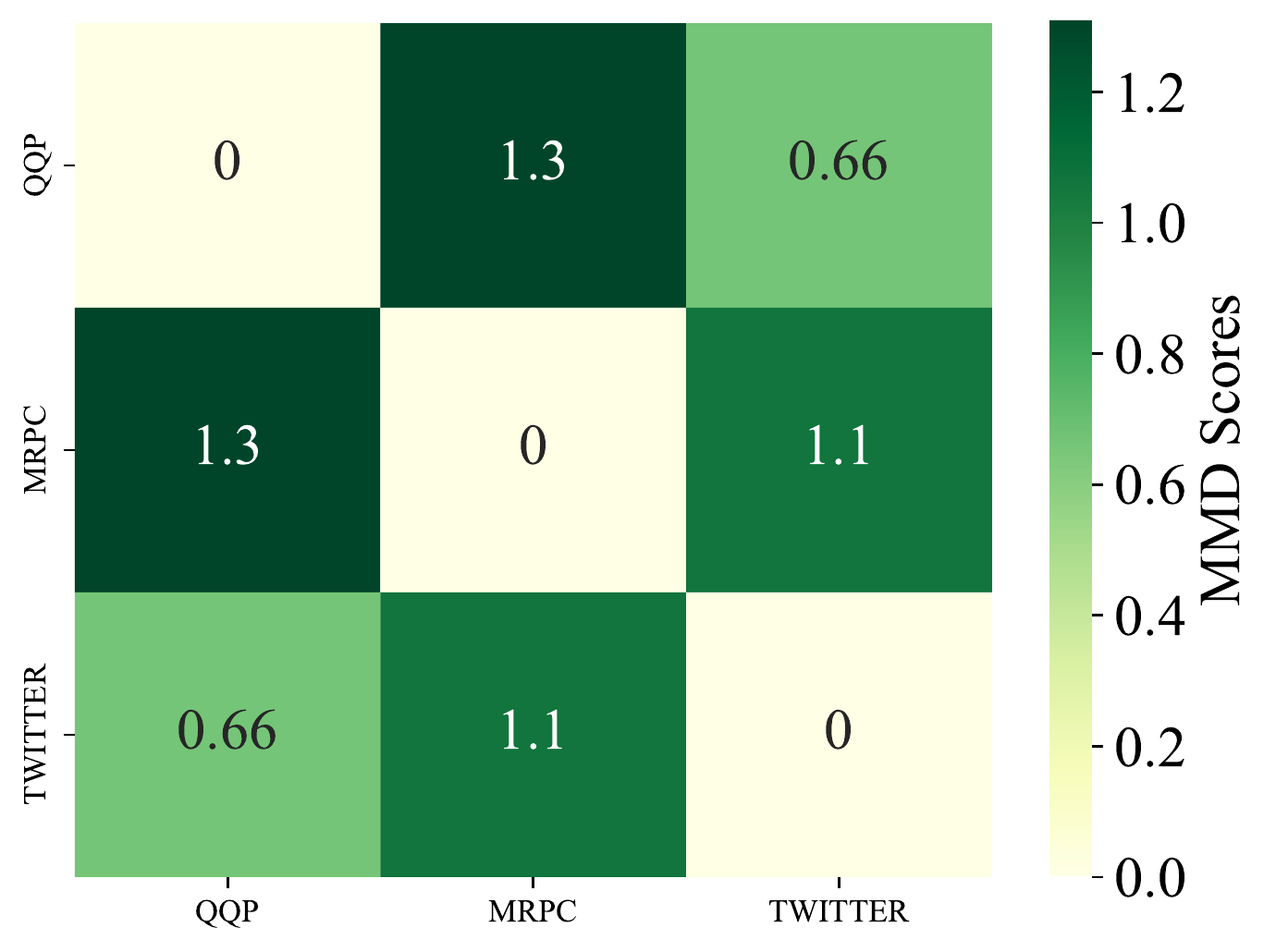}
\end{minipage}%
}%
                 
\subfigure[RTE]{
\begin{minipage}[t]{0.25\linewidth}
\centering
\includegraphics[width=.9\textwidth]{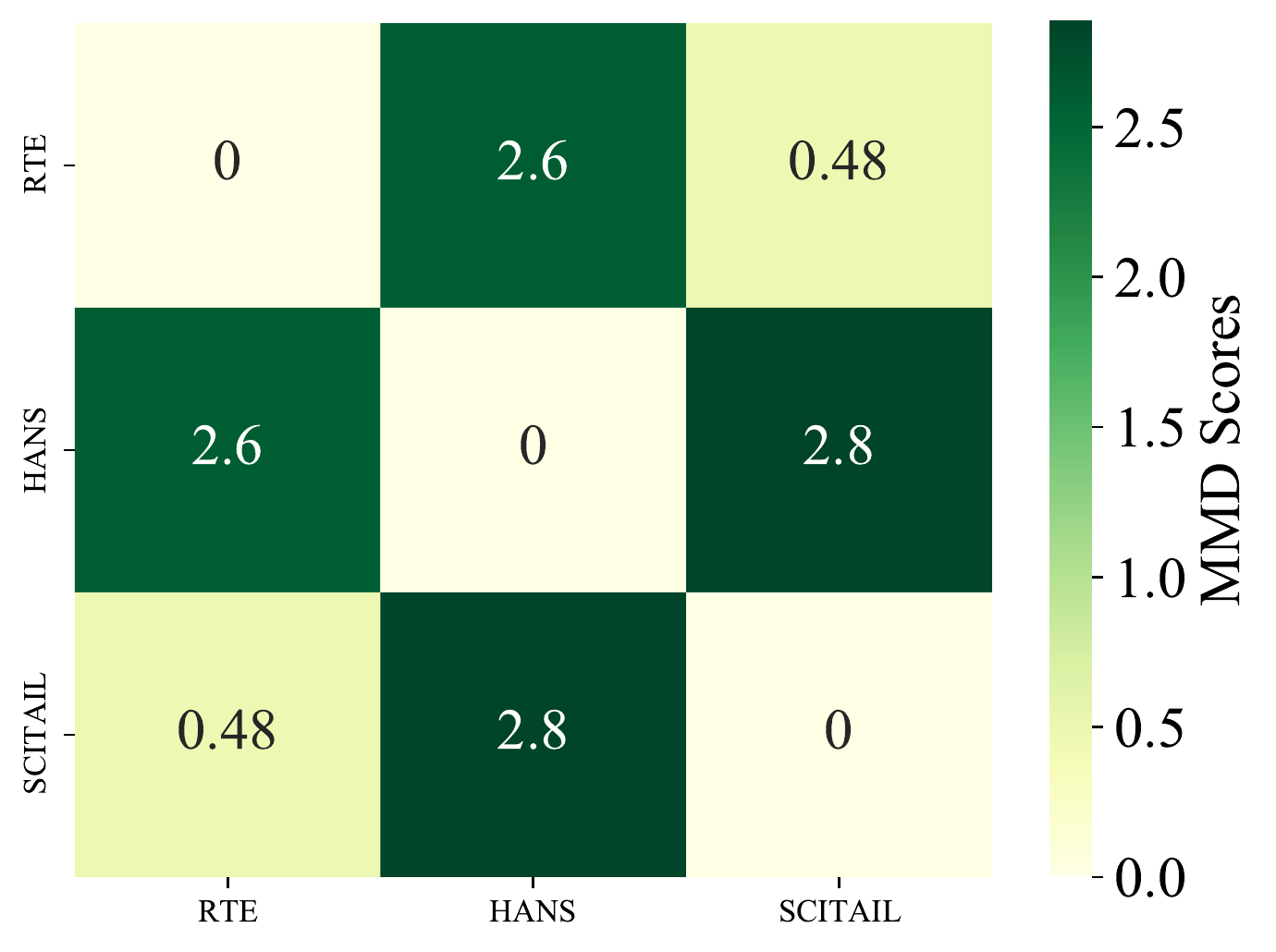}
\end{minipage}
}%
\subfigure[SST-2]{
\begin{minipage}[t]{0.25\linewidth}
\centering
\includegraphics[width=.9\textwidth]{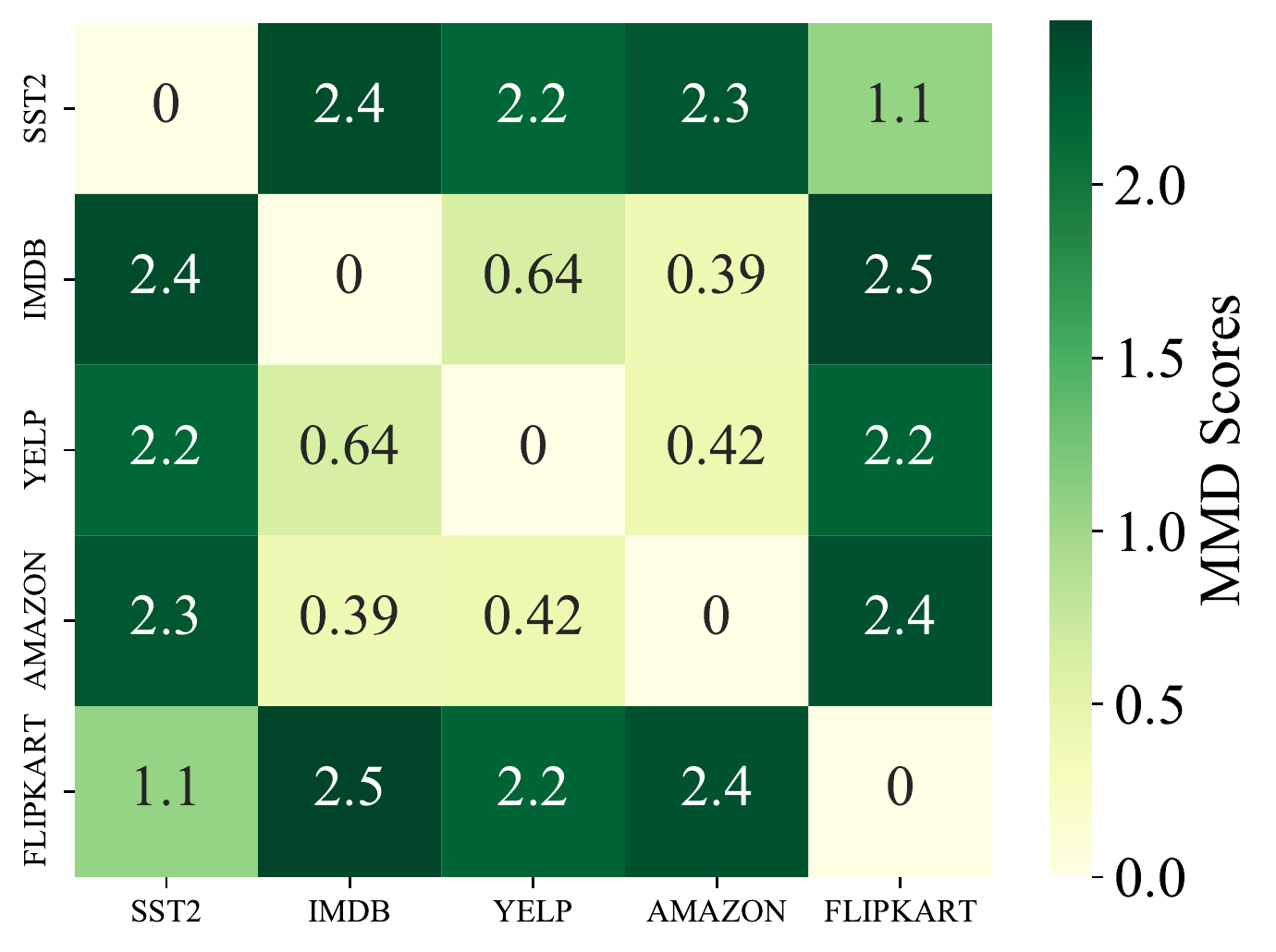}
\end{minipage}
}%
\subfigure[STSB]{
\begin{minipage}[t]{0.25\linewidth}
\centering
\includegraphics[width=.9\textwidth]{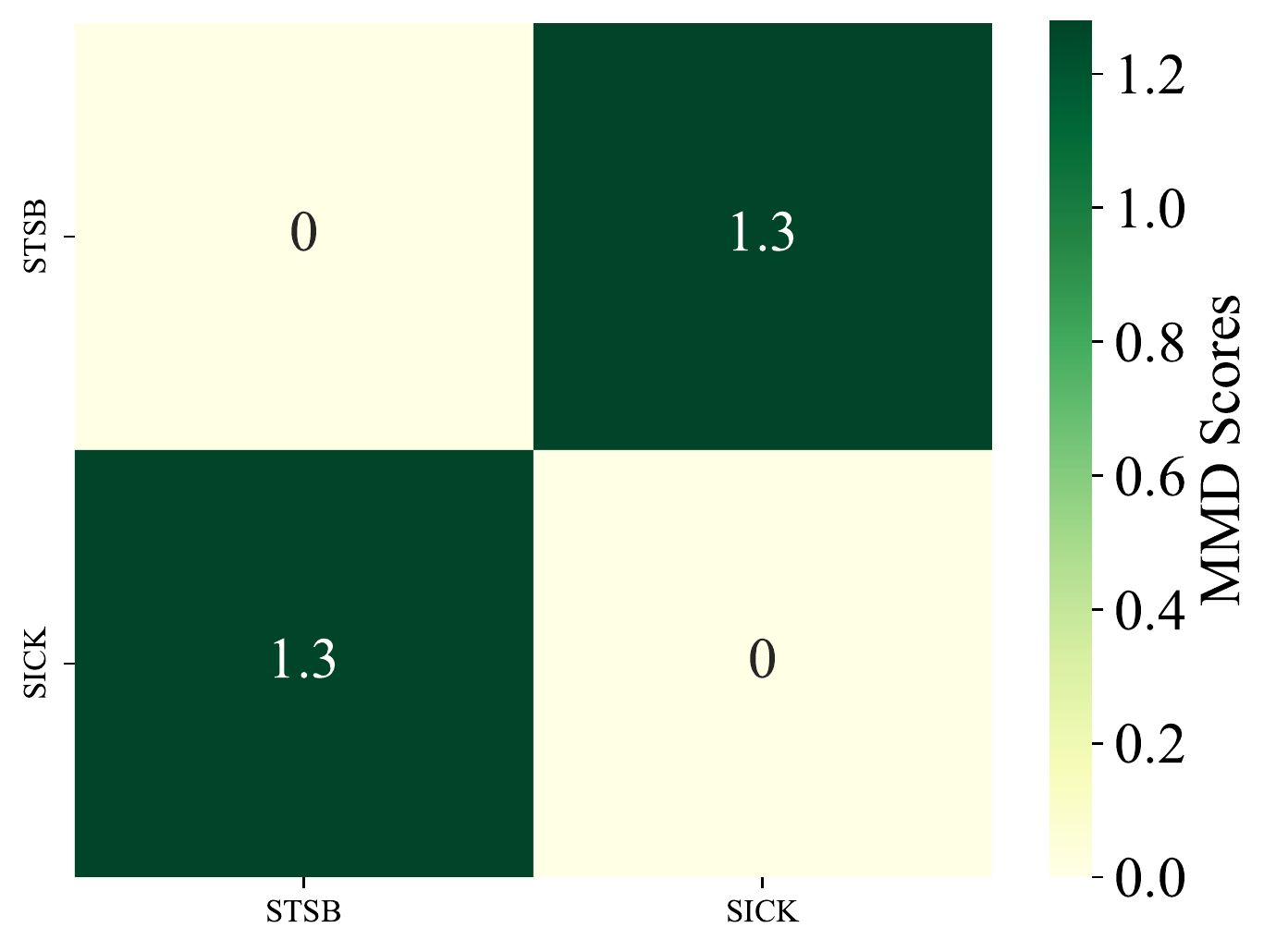}
\end{minipage}
}%
\subfigure[QNLI]{
\begin{minipage}[t]{0.25\linewidth}
\centering
\includegraphics[width=.9\textwidth]{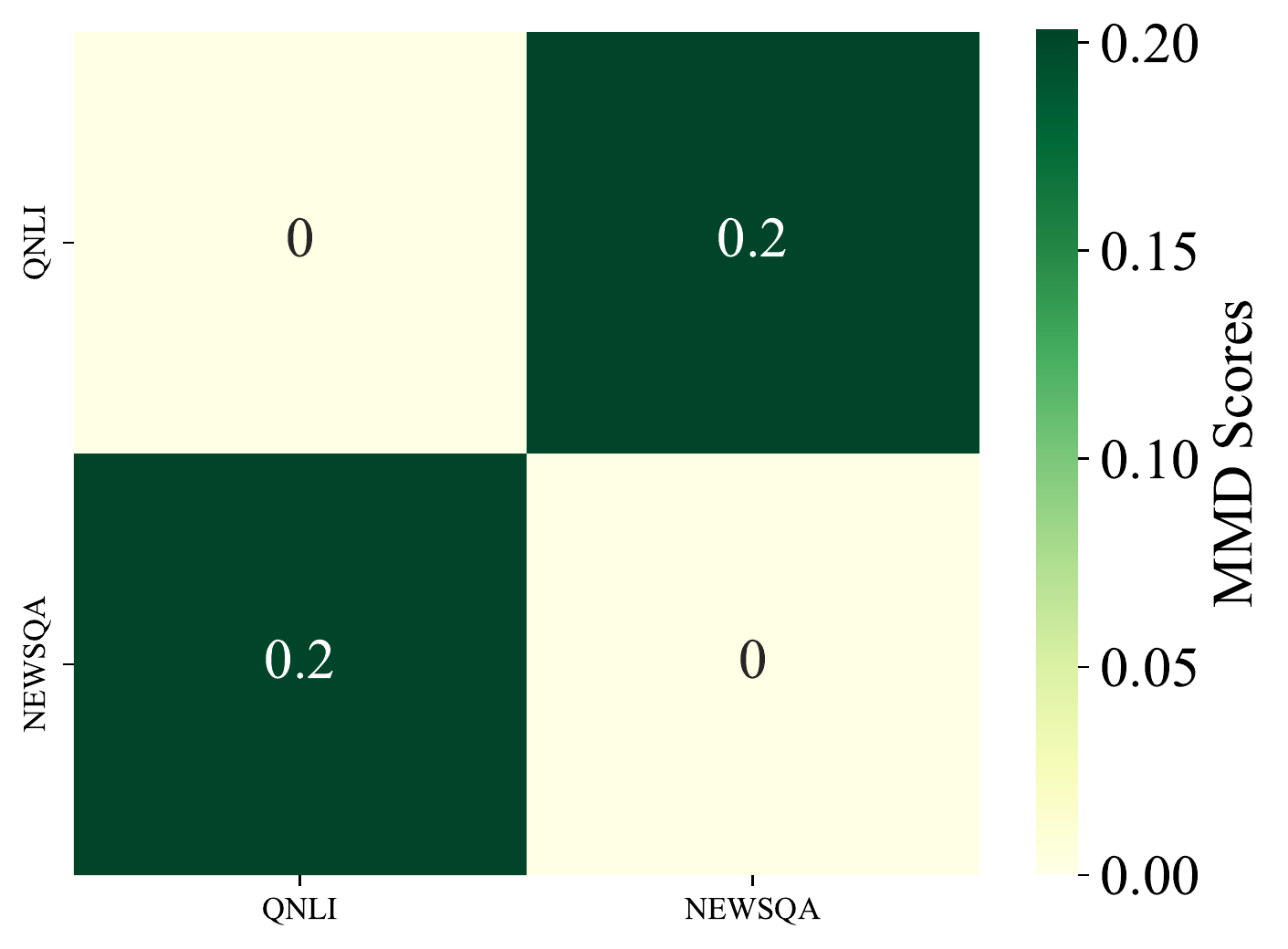}
\end{minipage}
}%
\centering
\caption{The MMD Scores between the training set and test set for each task. Lower MMD score means the higher correlation between datasets.}
\label{fig:MMD_scores}
\end{figure*}

\subsection{Word Overlap}

The similarity between datasets of In-distribution datasets and Out-of-distribution datasets are shown in Figure \ref{fig:Word_Overlap}.

\subsection{MMD Distance}
The MMD distance between ID and OOD datasets is shown in Figure \ref{fig:MMD_scores} for each task including in \method. When computing the MMD distance between two datasets, we ensure that the same number of sentences are sampled and fed into PLMs (e.g. RoBERTA-base) to extract their semantic features. We sample multiple times to get an average MMD sample score to estimate MMD distance of two datasets. The calculation of MMD is shown as follows:

\begin{equation}
\begin{aligned}
&\operatorname{MMD}^2[\mathcal{F}, X, Y]=\frac{1}{m(m-1)} \sum_{i \neq j}^m k\left(x_i, x_j\right) \\
&+\frac{1}{n(n-1)} \sum_{i \neq j}^n k\left(y_i, y_j\right)-\frac{2}{m n} \sum_{i, j=1}^{m, n} k\left(x_i, y_j\right)
\end{aligned}
\end{equation}

$\mathcal{F}$ is a MMD function class, \emph{i} and \emph{j} represents the batch of instances sampled from different distributions. \emph{m} and \emph{n} are the size of \emph{i} and \emph{j}.

\begin{table*}[t]
\centering
\small
\begin{adjustbox}{width=\textwidth}
\begin{tabular}{lcccccccccc}
\toprule
\multirow{2}{*}{\textbf{Model}} & \multicolumn{1}{c}{\textbf{SST-2}} & \multicolumn{1}{c}{\textbf{MNLI}} & \multicolumn{1}{c}{\textbf{QNLI}} & \multicolumn{1}{c}{\textbf{RTE}} & \multicolumn{1}{c}{\textbf{MRPC}} & \multicolumn{1}{c}{\textbf{QQP}} & \multicolumn{1}{c}{\textbf{STS-B}} & \multicolumn{1}{c}{\textbf{CoLA}} & \multicolumn{1}{c}{\textbf{Avg}} & \textbf{Avg}       \\
                       & \multicolumn{1}{c}{}   & \multicolumn{1}{c}{}  & \multicolumn{1}{c}{} & \multicolumn{1}{c}{}  & \multicolumn{1}{c}{} & \multicolumn{1}{c}{}   & \multicolumn{1}{c}{}  & \multicolumn{1}{c}{} & \multicolumn{1}{c}{} & $\Delta$↓ 
                       \\ \midrule
Humans (OOD) & \textit{92.36} & \textit{84.13} & \textit{81.10} & \textit{83.47} & \textit{84.70} & \textit{85.43} & \textit{80.28} & \textit{58.98} & \textit{80.14} & \textit{7.82}
                       \\ \midrule
GPT-3 (ID) & 93.68 & 69.27 & 79.20 & 80.20 & 79.21 & 72.15 & 88.10 & 50.13 & 76.49  & -  \\ 
GPT-3.5 (ID) & 95.75 & 72.25 & 82.78 & 82.71 & 73.36 & 75.69 & 89.55 & 54.99 & 78.39 & -  \\ 
\midrule
%

GPT-3 (OOD) & 92.33 & 61.50 & 79.00 & 71.03 & 59.55 & 55.41 & 73.74 & 27.31  & 64.98 & 11.51      \\ 
GPT-3.5 (OOD) & \textbf{95.92} & 66.01 & 75.84 & 66.15 & 58.43 & 67.96 & 74.01 & 30.77 & 66.90 & \textbf{11.49} \\
ELECTRA-large (OOD)& 95.14 & \textbf{76.94} & \textbf{80.44} & \textbf{78.74} & \textbf{69.96} & \textbf{77.24} & \textbf{81.14}  & \textbf{37.85} & \textbf{69.68}  & 21.87       \\ 
\bottomrule

\end{tabular}
\end{adjustbox}
\caption{OOD performance of GPT-3 and GPT3.5 using in-context learning compared with human performance and ELECTRA-large. We randomly select a single instance for each label. GPT-3 refers to text-davinci-003, and GPT-3.5 denotes the gpt-3.5-turbo.}
\label{tab:gpt3}
\end{table*}

\section{Rationale Overlap}
In order to measure the difference between rationales detected by PLMs and humans, we define precision as the percentage of the predicted rationales that also exist in the human annotation and recall as the percentage of words in the human annotation that also exist in the predicted rationales. We calculate the F1 score as an evaluation metric of overlap.

We show the evaluation of rationale overlap between models and humans on the e-SNLI dataset \cite{camburu2018snli} in Table \ref{tab:rationale_snli}. We find that the performance gap between different models is not very large (varying from 30.93 to 34.98). Models show a higher rationale overlap with humans based on e-SNLI than sentiment analysis datasets. This can be because the average length of instances in e-SNLI is generally shorter than that in sentiment analysis datasets. In particular, the base-sized ELECTRA has achieved the highest F1 score (34.98\%) among these models. 

\section{The In-domain Evaluation Results}

Following \cite{wang2018glue}, we report the in-domain evaluation results in Table \ref{fig:id_results}. We generally find that ELECTRA-large achieves the best average performance over seven tasks. Note that we report the results by evaluating models on the validation set provided by GLUE.

\section{The Correlation between ID and OOD Performance} \label{sec:appendixD}

\begin{figure*}[t]
\centering

\subfigure[MRPC]{
\begin{minipage}[t]{0.5\linewidth}
\centering
\includegraphics[width=.9\textwidth]{plots/ID_OOD/mrpc.pdf}
\end{minipage}%
}%
\subfigure[QQP]{
\begin{minipage}[t]{0.5\linewidth}
\centering
\includegraphics[width=.9\textwidth]{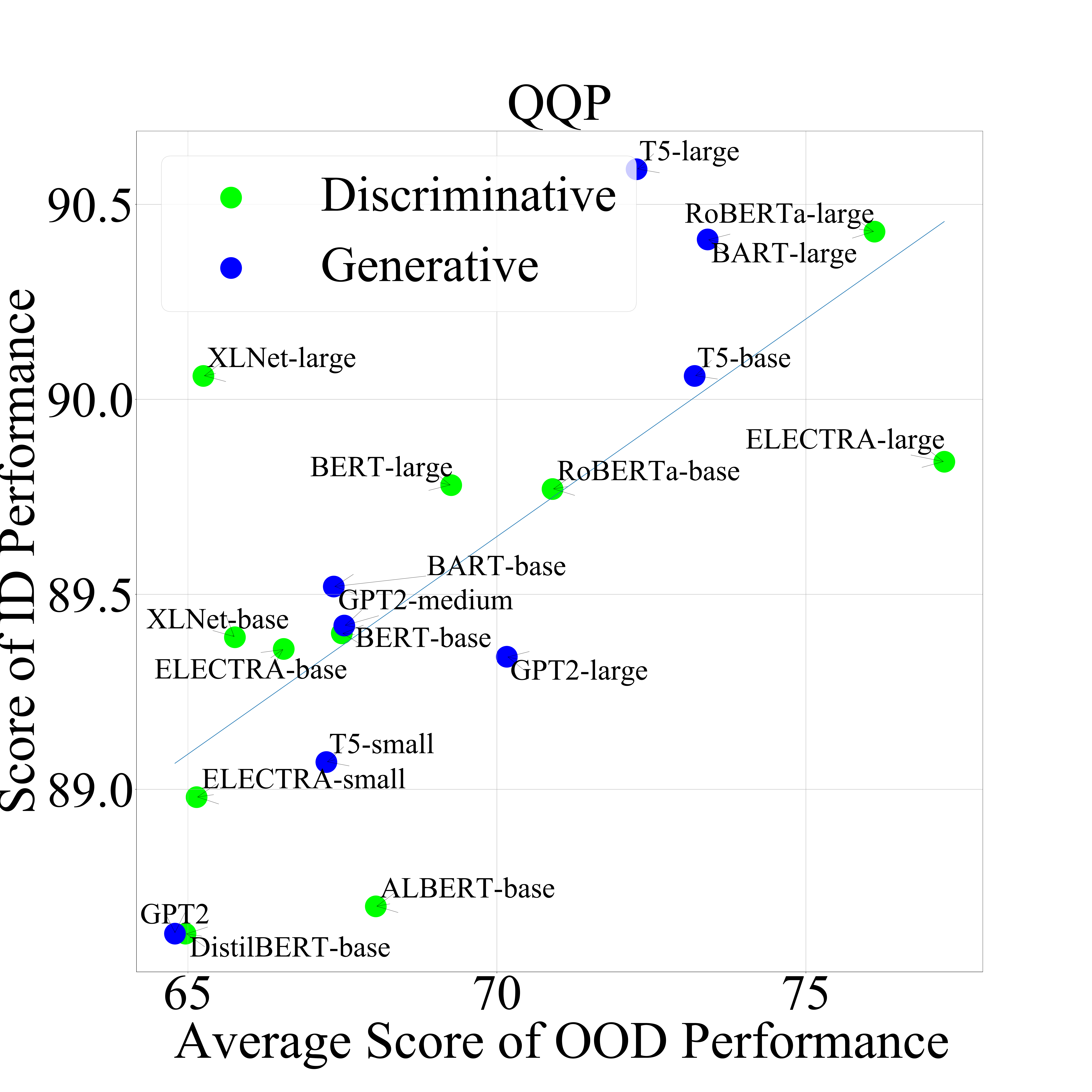}
\end{minipage}%
}%


\subfigure[STSB]{
\begin{minipage}[t]{0.5\linewidth}
\centering
\includegraphics[width=.9\textwidth]{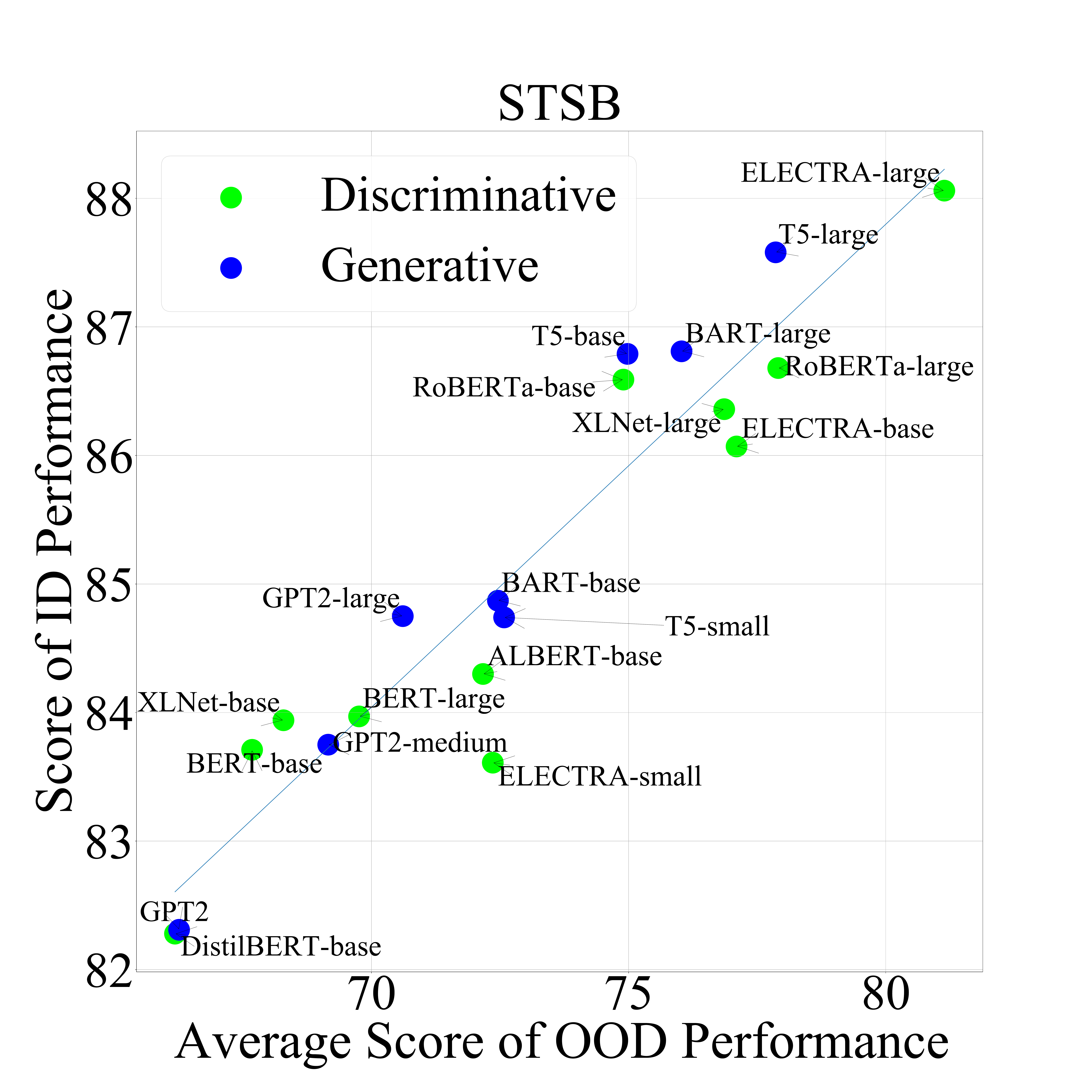}
\end{minipage}
}%
\subfigure[QNLI]{
\begin{minipage}[t]{0.5\linewidth}
\centering
\includegraphics[width=.9\textwidth]{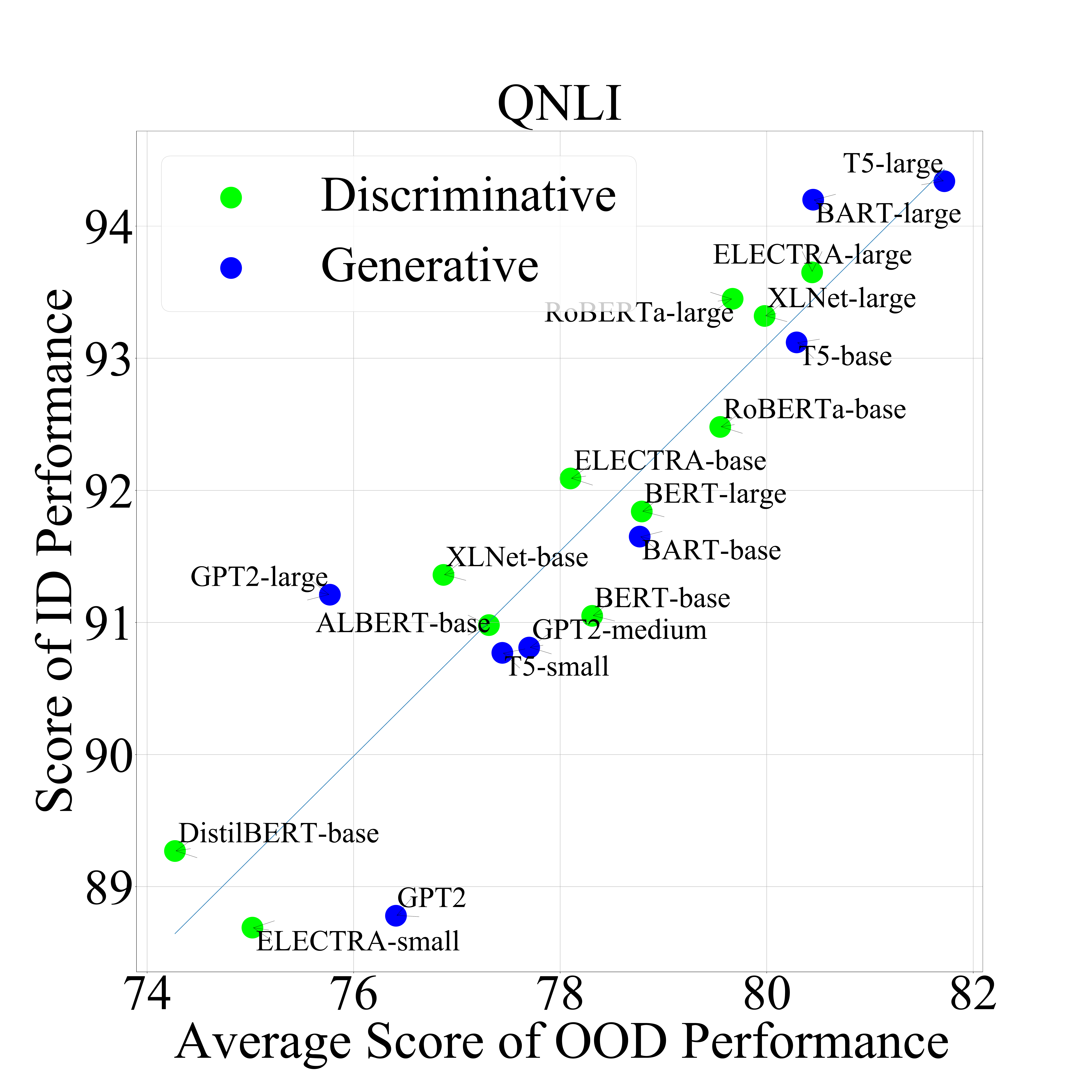}
\end{minipage}
}%

\subfigure[RTE]{
\begin{minipage}[t]{0.5\linewidth}
\centering
\includegraphics[width=.9\textwidth]{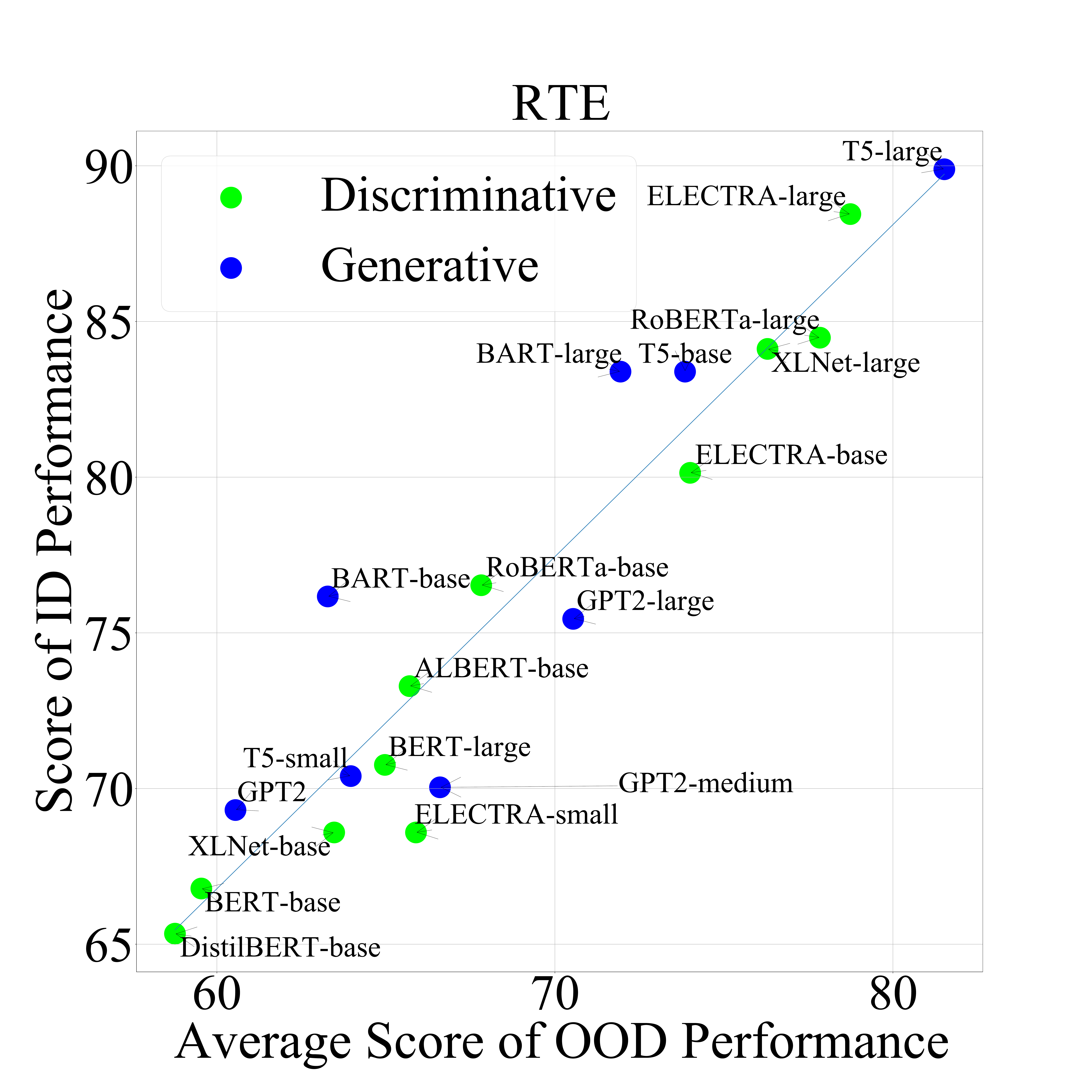}
\end{minipage}%
}%
\subfigure[SST2]{
\begin{minipage}[t]{0.5\linewidth}
\centering
\includegraphics[width=.9\textwidth]{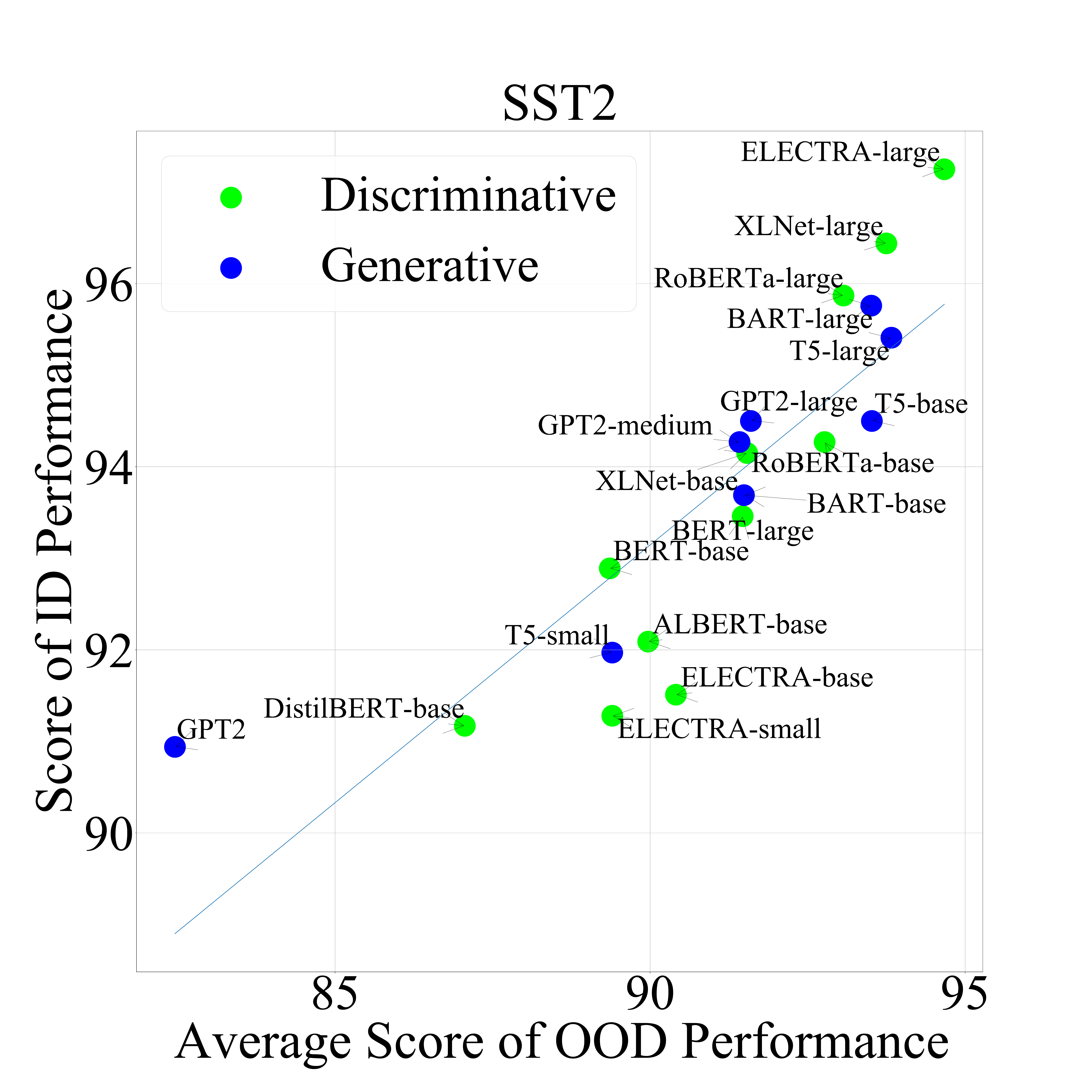}
\end{minipage}%
}%


\centering
\caption{The correlation between the ID and OOD performance for each task involving in \method.}
\label{fig:scatter_appendix}
\end{figure*}

In general, we find that the overall performance of ID and OOD tests shows a linear correlation for both discriminative and generative models. In addition to the overall performance, we look at task-level performance at a more granular level in Figure \ref{fig:scatter_appendix}. As shown in Figure \ref{fig:scatter_appendix}, we find that the linear correlation does not exist for every task. For example, the  linear correlation is extremely weak for MRPC and QQP, with relatively low OOD accuracy. While the linear correlation becomes significant on STSB and QNLI.

\section{The ID and OOD Performance of GPT-3 and GPT-3.5} \label{sec:appendixG}

\textbf{Settings.} The performance of GPT-3 and GPT-3.5 is shown in Table \ref{tab:gpt3}, where we report the classification results based on 1,000 instances for each task. The training strategy of GPT-3 is simulated to keep the same as human evaluation. We feed the model with some in-domain instances as instructions before testing on the OOD dataset. To achieve this, we adopt the official API for calculating the in-domain performance of GPT-3 (text-DaVinci-003) based on 1,000 sampled ID instances. We leverage the in-context learning following \cite{ouyang2022training} to calculate its OOD results on \method.

\noindent\textbf{Results.} In Table \ref{tab:gpt3}, it is interesting to see that the performance decay ratio of GPT-3 caused by the domain generalization is similar to GPT-3.5 while significantly larger than Humans (11.49\% -- GPT 3.5 vs. 11.51\% -- GPT-3 vs. 7.82\% -- Humans), indicating that there is much room for improvement in the OOD robustness, even for state-of-the-art LLMs. Meanwhile, it can be seen that the OOD performance of GPT-3.5 is still far behind the human performance (66.90\% -- GPT-3 vs. 80.14\% -- Humans), and slightly lower than ELECTRA-large (69.68\%). Notably, the results of GPT-3/3.5 should be \textbf{treated with caution and just for reference} because we are not sure if datasets of \method are already included in the training corpus of GPT-3/3.5. Also, the OOD performance listed in Table \ref{tab:gpt3} cannot be compared with PLMs fairly, as we only adopt instructions to evaluate it not fine-tuning the model like other PLMs in \method.

\begin{table*}[t]
\centering
\small
\begin{tabular}{lllllllllll}
\hline
Model   & \multicolumn{1}{l}{SST-2} & \multicolumn{1}{l}{MNLI} & \multicolumn{1}{l}{QNLI} &\multicolumn{1}{l}{RTE} & \multicolumn{1}{l}{MRPC} & \multicolumn{1}{l}{QQP} & \multicolumn{1}{l}{STSB} & \multicolumn{1}{l}{COLA} & \multicolumn{1}{l}{Average} & \multicolumn{1}{l}{Parameters}  \\ \hline
ELECTRA-large   & 97.25 & 89.29 & 93.65 & 88.45 & 92.60  & 89.84 & 88.06 & 74.33 & 89.18              & 334.09                          \\
RoBERTa-large   & 95.87 & 89.47 & 93.45 & 84.48 & 92.36 & 90.43 & 86.68 & 69.90  & 87.83              & 355.36                          \\
T5-large        & 95.41 & 88.83 & 94.34 & 89.89 & 92.01 & 90.59 & 87.58 & 62.97 & 87.70               & 737.67                          \\
BART-large      & 95.76 & 88.30  & 94.20  & 83.39 & 92.21 & 90.41 & 86.81 & 65.29 & 87.05              & 406.29                          \\
XLNet-large     & 96.44 & 89.50  & 93.32 & 84.12 & 91.54 & 90.06 & 86.36 & 62.63 & 86.75              & 360.27                          \\
T5-base         & 94.50  & 86.55 & 93.12 & 83.39 & 91.22 & 90.06 & 86.79 & 61.71 & 85.92              & 222.9                           \\
ELECTRA-base    & 91.51 & 87.12 & 92.09 & 80.14 & 91.09 & 89.36 & 86.07 & 69.95 & 85.92              & 108.89                          \\
RoBERTa-base    & 94.27 & 87.43 & 92.48 & 76.53 & 91.83 & 89.77 & 86.59 & 63.25 & 85.27              & 124.65                          \\
GPT2-large      & 94.50  & 85.48 & 91.21 & 75.45 & 87.78 & 89.34 & 84.75 & 60.06 & 83.57              & 774.03                          \\
BERT-large      & 93.46 & 85.69 & 91.84 & 70.76 & 90.26 & 89.78 & 83.97 & 60.32 & 83.26              & 335.14                          \\
BART-base       & 93.69 & 85.89 & 91.65 & 76.17 & 89.75 & 89.52 & 84.87 & 52.78 & 83.04              & 139.42                          \\
ALBERT-base     & 92.09 & 83.81 & 90.98 & 73.29 & 90.23 & 88.70  & 84.30  & 57.25 & 82.58              & 11.68                           \\
XLNet-base      & 94.15 & 86.49 & 91.36 & 68.59 & 90.50  & 89.39 & 83.94 & 53.67 & 82.26              & 116.72                          \\
BERT-base       & 92.89 & 83.63 & 91.05 & 66.79 & 89.41 & 89.40  & 83.71 & 59.75 & 82.08              & 109.48                          \\
GPT2-medium     & 94.27 & 85.38 & 90.81 & 70.04 & 87.20  & 89.42 & 83.75 & 53.87 & 81.84              & 354.82                          \\
ELECTRA-small   & 91.28 & 81.93 & 88.69 & 68.59 & 89.88 & 88.98 & 83.61 & 59.06 & 81.50               & 13.48                           \\
T5-small        & 91.97 & 82.82 & 90.77 & 70.40  & 89.13 & 89.07 & 84.74 & 43.88 & 80.35              & 60.51                           \\
DistilBERT-base & 91.17 & 82.20  & 89.27 & 65.34 & 88.33 & 88.63 & 82.28 & 54.43 & 80.21              & 66.36                           \\
GPT2            & 90.94 & 82.63 & 88.78 & 69.31 & 84.51 & 88.63 & 82.31 & 47.29 & 79.30               & 124.44                     \\     \hline
\end{tabular}
\caption{Detailed results of the in-domain test on each task sorted by the average performance.}
\label{fig:id_results}
\end{table*}


\begin{table*}[ht]
\centering
\small
\begin{tabular}{lllllllll}
\cline{1-9}
 Model           & SST2     & MNLI     & QNLI     & RTE      & MRPC     & QQP      & STSB     & COLA      \\ \hline
ELECTRA-large   & 2e-05/64 & 2e-05/16 & 5e-05/64 & 5e-05/32 & 2e-05/16 & 2e-05/64 & 2e-05/64 & 2e-05/128 \\ 
RoBERTa-large   & 2e-05/32 & 2e-05/64 & 2e-05/32 & 2e-05/32 & 2e-05/16 & 2e-05/64 & 2e-05/16 & 2e-05/32  \\
T5-large        & 1e-4/16  & 1e-4/32  & 1e-4/32  & 1e-4/32  & 1e-4/32  & 1e-4/16  & 1e-4/64  & 1e-4/32   \\
BART-large      & 2e-05/32 & 2e-05/16 & 2e-05/32 & 3e-05/32 & 2e-05/32 & 2e-05/32 & 2e-05/30 & 2e-05/32  \\
XLNet-large     & 3e-05/64 & 2e-05/64 & 3e-05/32 & 1e-05/32 & 1e-05/16 & 2e-05/32 & 2e-05/16 & 2e-05/16  \\
T5-base         & 1e-4/32  & 1e-4/16  & 1e-4/32  & 1e-4/8   & 1e-4/16  & 1e-4/16  & 3e-4/16  & 1e-4/32   \\
ELECTRA-base    & 1e-4/32  & 5e-05/64 & 5e-05/64 & 5e-05/16 & 5e-05/16 & 5e-05/32 & 5e-05/16 & 5e-05/32  \\
RoBERTa-base    & 2e-05/32 & 2e-05/32 & 2e-05/32 & 2e-05/32 & 3e-05/32 & 2e-05/32 & 3e-05/32 & 2e-05/16  \\
GPT2-large      & 2e-05/32 & 2e-05/32 & 2e-05/32 & 2e-05/16 & 3e-05/32 & 2e-05/32 & 2e-05/32 & 2e-05/32  \\
BERT-large      & 2e-05/32 & 2e-05/32 & 2e-05/16 & 2e-05/16 & 2e-05/16 & 2e-05/64 & 2e-05/64 & 3e-05/16  \\
BART-base       & 2e-05/32 & 2e-05/32 & 2e-05/32 & 2e-05/16 & 3e-05/32 & 2e-05/16 & 2e-05/16 & 2e-05/32  \\
ALBERT-base     & 2e-05/32 & 2e-05/32 & 2e-05/32 & 2e-05/32 & 2e-05/16 & 2e-05/32 & 2e-05/32 & 2e-05/32  \\
XLNet-base      & 3e-05/32 & 2e-05/32 & 2e-05/32 & 1e-05/16 & 2e-05/16 & 2e-05/32 & 2e-05/32 & 1e-05/32  \\
BERT-base       & 2e-05/32 & 3e-05/32 & 2e-05/32 & 3e-05/32 & 3e-05/32 & 2e-05/32 & 2e-05/16 & 3e-05/32  \\
GPT2-medium     & 2e-05/32 & 2e-05/16 & 3e-05/32 & 3e-05/32 & 3e-05/16 & 3e-05/32 & 3e-05/32 & 3e-05/32  \\
ELECTRA-small   & 5e-05/64 & 5e-05/64 & 5e-05/32 & 5e-05/64 & 5e-05/32 & 5e-05/32 & 5e-05/32 & 5e-05/64  \\
T5-small        & 1e-4/16  & 1e-4/16  & 1e-4/32  & 3e-4/16  & 1e-4/16  & 1e-4/16  & 3e-4/32  & 3e-4/32   \\
DistilBERT-base & 3e-05/16 & 2e-05/32 & 3e-05/32 & 2e-05/16 & 2e-05/16 & 2e-05/16 & 2e-05/16 & 2e-05/16  \\
GPT2            & 2e-05/32 & 2e-05/32 & 3e-05/32 & 2e-05/32 & 2e-05/32 & 2e-05/32 & 3e-05/32 & 3e-05/32 \\
\hline \cline{1-9}
\end{tabular}
\caption{The hyper-parameter setting for each task, including the learning rate and batch size.}
\label{tab:Hyperparameters}
\end{table*}

\end{document}